\documentclass[]{article}
\pdfoutput=1
\newcommand{\ArxivUai}[2]{#1}

\usepackage{proceed2e}
\usepackage{times}
\usepackage{graphicx}
\usepackage{amsthm}
\usepackage[figure, noline]{algorithm2e}

\newtheorem{definition}{Definition}
\newtheorem{theorem}{Theorem}
\newtheorem{corollary}{Corollary}
\newtheorem{proposition}{Proposition}
\newtheorem{lemma}{Lemma}

\newcommand{\lessequiv}{\leq}
\newcommand{\Gr}{{\cal G}}
\newcommand{\gr}{{\cal G}}
\newcommand{\Hr}{{\cal H}} 

\newcommand{\Parset}[2]{\mbox{${\bf Pa}^{#1}_{#2}$}}

\newcommand{\NonDesc}[2]{\mbox{${\bf NonDe}^{#1}_{#2}$}}
\newcommand{\indep}{\mbox{$\perp\!\!\perp$}}
\newcommand{\notindep}{\mbox{$\not \!\!\! \indep$}}
\newcommand{\SSet}{\mbox{${\bf S}$}}

\newcommand{\HSet}{\mbox{${\bf H}$}}

\newcommand{\Data}{\mbox{${\bf D}$}}

\newcommand{\Parnot}[2]{\mbox{${\bf Pa}^{\Hr}_{#1\downarrow#2}$}}
\newcommand{\gleaf}{\mbox{$\Gr$-leaf}\ }

\newcommand{\NeiAdj}[2]{{\bf NA}_{#1,#2}}
\newcommand{\pdag}{\mbox{$\cal P$}}
\newcommand{\qdag}{\mbox{$\cal Q$}}
\newcommand{\cpdag}{\mbox{${\cal C}$}}

\newcommand{\prune}[2]{\mbox{${\bf Prune}(#1, #2)$}}
\newcommand{\prunegh}{\prune{\Gr}{\Hr}}

\newcommand{\bV}{\mbox{${\bf V}$}}
\newcommand{\bS}{\mbox{${\bf S}$}}
\newcommand{\bR}{\mbox{${\bf R}$}}
\newcommand{\bT}{\mbox{${\bf T}$}}
\newcommand{\bD}{\mbox{${\bf D}$}}

\newcommand{\bC}{\mbox{${\bf C}$}}

\newcommand{\bX}{\mbox{${\bf X}$}}
\newcommand{\bY}{\mbox{${\bf Y}$}}
\newcommand{\bZ}{\mbox{${\bf Z}$}}
\newcommand{\bW}{\mbox{${\bf W}$}}

\newcommand{\Eqclass}{\mbox{$\cal E$}}
\newcommand{\Fqclass}{\mbox{$\cal F$}}

\newcommand{\Thetasetval}{\mbox{\boldmath $\theta$}}

\newcommand{\EClass}[1]{\mbox{$[#1]_{\approx}$}}
\newcommand{\ISG}[2]{\mbox{$#1[#2]$}}

\newcommand{\cindArg}[4]{\mbox{$#1 \indep_{#4} #2 | #3$}}

\newcommand{\cind}[3]{\cindArg{#1}{#2}{#3}{}}
\newcommand{\gcind}[3]{\cindArg{#1}{#2}{#3}{\mbox{$\Gr$}}}
\newcommand{\hcind}[3]{\cindArg{#1}{#2}{#3}{\mbox{$\Hr$}}}

\newcommand{\axand}{\hspace{0.05in}{+}\hspace{0.1in}}
\newcommand{\axor}{\hspace{0.05in}{OR}\hspace{0.1in}}

\newcommand{\comment}[1]{}

\title{Selective Greedy Equivalence Search: Finding Optimal Bayesian
  Networks Using a Polynomial Number of Score Evaluations}

\author{{\bf David Maxwell Chickering} \\
Microsoft Research\\
Redmond, WA 98052\\
dmax@microsoft.com\\
\And
{\bf Christopher Meek} \\
Microsoft Research\\
Redmond, WA 98052\\
meek@microsoft.com\\
}

\begin{document}
\maketitle


\begin{abstract}
We introduce Selective Greedy Equivalence Search (SGES), a restricted
version of Greedy Equivalence Search (GES). SGES retains the
asymptotic correctness of GES but, unlike GES, has polynomial
performance guarantees. In particular, we show that when data are
sampled independently from a distribution that is perfect with respect
to a DAG $\Gr$ defined over the observable variables then, in the
limit of large data, SGES will identify $\Gr$'s equivalence class
after a number of score evaluations that is (1) polynomial in the
number of nodes and (2) exponential in various complexity measures
including maximum-number-of-parents, maximum-clique-size, and a new
measure called {\em v-width} that is at least as small as---and
potentially much smaller than---the other two.  More generally, we
show that for any hereditary and equivalence-invariant property $\Pi$
known to hold in $\Gr$, we retain the large-sample optimality
guarantees of GES even if we ignore any GES deletion operator during
the backward phase that results in a state for which $\Pi$ does not
hold in the common-descendants subgraph.
\end{abstract}

\section{INTRODUCTION}

Greedy Equivalence Search (GES) is a score-based search algorithm that
searches over equivalence classes of Bayesian-network structures. The
algorithm is appealing because (1) for finite data, it explicitly (and
greedily) tries to maximize the score of interest, and (2) as the data
grows large, it is guaranteed---under suitable distributional
assumptions---to return the generative structure. Although
empirical results show that the algorithm is efficient in real-world
domains, the number of search states that GES needs to evaluate in the
worst case can be exponential in the number of domain variables.

In this paper, we show that if we assume the generative distribution
is perfect with respect to some DAG $\Gr$ defined over the observable
variables, and if $\Gr$ is known to be constrained by various
graph-theoretic measures of complexity, then we can disregard all but
a polynomial number of the backward search operators considered by GES
while retaining the large-sample guarantees of the algorithm; we call
this new variant of GES {\em selective greedy equivalence search} or
{\em SGES}. Our complexity results are a consequence of a new
understanding of the backward phase of GES, in which edges (either
directed or undirected) are greedily deleted from the current state
until a local minimum is reached. We show that for any {\em
  hereditary} and {\em equivalence-invariant} property known to hold
in generative model $\Gr$, we can remove from consideration any
edge-deletion operator between $X$ and $Y$ for which the property does
not hold in the resulting induced subgraph over $X,$ $Y,$ and their
common descendants. As an example, if we know that each node has at
most $k$ parents, we can remove from consideration any deletion
operator that results in a common child with more than $k$ parents.

We define a new notion of complexity that we call {\em v-width}.  For
a given generative structure $\Gr$, v-width is necessarily smaller
than the maximum clique size, which is necessarily smaller than or
equal to the maximum number of parents per node.  By casting limited
v-width and other complexity constraints as graph properties, we show
how to enumerate directly over a polynomial number of edge-deletion
operators at each step, and we show that we need only a polynomial
number of calls to the scoring function to complete the algorithm.

The main contributions of this paper are theoretical. Our definition
of the new SGES algorithm deliberately leaves unspecified the details
of how to implement its forward phase; we prove our results for SGES
given {\em any} implementation of this phase that completes with a
polynomial number of calls to the scoring function. A naive
implementation is to immediately return a complete (i.e., no
independence) graph using {\em no} calls to the scoring function, but
this choice is unlikely to be reasonable in practice, particularly in
discrete domains where the sample complexity of this initial model
will likely be a problem. Whereas we believe it an important
direction, our paper does not explore practical alternatives for the
forward phase that have polynomial-time guarantees.

\ArxivUai{
This paper, which is an expanded version of Chickering and Meek (2015)
and includes all proofs, is organized as
follows. \nocite{ChickeringMeek2015UAI}
}{
Our paper is organized as follows.  }  In Section \ref{sec:related},
we describe related work. In Section \ref{sec:notation}, we provide
notation and background material. In Section \ref{sec:main}, we
present our new SGES algorithm, we show that it is optimal in the
large-sample limit, and we provide complexity bounds when given an
equivalence-invariant and hereditary property that holds on the
generative structure. In Section \ref{sec:experiments}, we present a
simple synthetic experiment that demonstrates the value of restricting
the backward operators in SGES. We conclude with a discussion of our
results in Section \ref{sec:conclude}.

\section{RELATED WORK} \label{sec:related}

It is useful to distinguish between approaches to learning the
structure of graphical models as {\em constraint based}, {\em score
  based} or {\em hybrid}. Constraint-based approaches typically use
(conditional) independence tests to eliminate potential models,
whereas score-based approaches typically use a penalized likelihood or
a marginal likelihood to evaluate alternative model structures; hybrid
methods combine these two approaches. Because score-based approaches
are driven by a global likelihood, they are less susceptible than
constraint-based approaches to incorrect categorical decisions about
independences.

There are polynomial-time algorithms for learning the best model in
which each node has at most one parent. In particular, the Chow-Liu
algorithm (Chow and Liu, 1968) used with any equivalence-invariant
score will identify the highest-scoring tree-like model in polynomial
time; for scores that are not equivalence invariant, we can use the
polynomial-time maximum-branching algorithm of Edmonds (1967)
instead. Gaspers et al. (2012) show how to learn {\em k-branchings} in
polynomial time; these models are polytrees that differ from a
branching by a constant $k$ number of edge deletions.

\nocite{Chow68}
\nocite{Edmonds67}
\nocite{Gaspers2012}

Without additional assumptions, most results for learning
non-tree-like models are negative. Meek (2001) shows that finding the
maximum-likelihood path is NP-hard, despite this being a special case
of a tree-like model. Dasgupta (1999) shows that finding the
maximum-likelihood polytree (a graph in which each pair of nodes is
connected by at most one path) is NP-hard, even with bounded in-degree
for every node. For general directed acyclic graphs, Chickering (1996)
shows that finding the highest marginal-likelihood structure under a
particular prior is NP-hard, even when each node has at most two
parents. Chickering at al. (2004) extend this same result to the
large-sample case.

\nocite{Meek2001}
\nocite{Dasgupta99}
\nocite{Chickering96lns}
\nocite{CMH2004}

Researchers often assume that the training-data ``generative''
distribution is {\em perfect} with respect to some model class in
order to reduce the complexity of learning algorithms. Geiger et
al. (1990) provide a polynomial-time constraint-based algorithm for
recovering a polytree under the assumption that the generative
distribution is perfect with respect to a polytree; an analogous
score-based result follows from this paper. The constraint-based PC
algorithm of Sprites et al. (1993) can identify the equivalence class
of Bayesian networks in polynomial time if the generative structure is
a DAG model over the observable variables in which each node has a
bounded degree; this paper provides a similar result for a score-based
algorithm.  Kalish and Buhlmann (2007) show that for Gaussian
distributions, the PC algorithm can identify the right structure even
when the number of nodes in the domain is larger than the sample size.
Chickering (2002) uses the same DAG-perfectness-over-observables
assumption to show that the greedy GES algorithm is optimal in the
large-sample limit, although the branching factor of GES is worst-case
exponential; the main result of this paper shows how to limit this
branching factor without losing the large-sample guarantee.
Chickering and Meek (2002) show that GES identifies a ``minimal''
model in the large-sample limit under a less restrictive set of
assumptions.

\nocite{GPP1990}
\nocite{Spirtes93}
\nocite{Kalisch2007}
\nocite{Chickering02JMLRb}
\nocite{CM02uai}

Hybrid methods for learning DAG models use a constraint-based
algorithm to prune out a large portion of the search space, and then
use a score-based algorithm to select among the remaining (Friedman et
al., 1999; Tsamardinos et al., 2006). Ordyniak and Szeider (2013) give
positive complexity results for the case when the remaining DAGs are
characterized by a structure with constant treewidth.

\nocite{Friedman1999uai}
\nocite{Tsamardinos2006ML}
\nocite{Ordyniak2013}

Many researchers have turned to exhaustive enumeration to identify the
highest-scoring model (Gillispie and Perlman, 2001; Koivisto and Sood
2004; Silander and Myllym\"{a}ki, 2006; 
Kojima et al, 2010). There are many complexity results for other model
classes. Karger and Srebro (2001) show that finding the optimal Markov
network is NP-complete for treewidth $> 1$. Narasimhan and Bilmes
(2004) and Shahaf, Chechetka and Guestrin (2009) show how to learn
approximate limited-treewidth models in polynomial time. Abeel, Koller
and Ng (2005) show how to learn factor graphs in polynomial
time.

\nocite{GP01}
\nocite{KoivistoSood2004}
\nocite{Silander2006}
\nocite{Shahaf2009}
\nocite{Abbeel2006}
\nocite{NarasimhanBilmes2004uai}
\nocite{Kojima2010}
\nocite{KargerSrebro2001}

\section{NOTATION AND BACKGROUND} \label{sec:notation}

We use the following syntactical conventions in this paper. We denote
a variable by an upper case letter (e.g., $A$) and a state or value of
that variable by the same letter in lower case (e.g., $a$). We denote
a set of variables by a bold-face capitalized letter or letters (e.g.,
${\bf X}$). We use a corresponding bold-face lower-case letter or
letters (e.g., ${\bf x}$) to denote an assignment of state or value to
each variable in a given set.  We use calligraphic letters (e.g.,
$\gr, \Eqclass$) to denote statistical models and graphs.

A {\em Bayesian-network model} for a set of variables ${\bf U}$ is a
pair $(\gr, \Thetasetval)$. $\gr = ({\bf V, E})$ is a directed acyclic
graph---or {\em DAG} for short---consisting of nodes in one-to-one
correspondence with the variables and directed edges that connect
those nodes. $\Thetasetval$ is a set of parameter values that specify
all of the conditional probability distributions. The Bayesian network
represents a joint distribution over ${\bf U}$ that factors according
to the structure $\gr$.  

The structure $\gr$ of a Bayesian network represents the independence
constraints that must hold in the distribution.  The set of all
independence constraints implied by the structure $\gr$ can be
characterized by the {\em Markov conditions}, which are the
constraints that each variable is independent of its non-descendants
given its parents. All other independence constraints follow from
properties of independence.  A distribution defined over the variables
from $\Gr$ is {\em perfect with respect to $\Gr$} if the set of
independences in the distribution is equal to the set of independences
implied by the structure $\Gr$.

Two DAGs $\gr$ and $\gr'$ are {\em equivalent}\footnote{We make the
  standard conditional-distribution assumptions of multinomials for
  discrete variables and Gaussians for continuous variables so that if
  two DAGs have the same independence constraints, then they can also
  model the same set of distributions.}---denoted $\Gr \approx
\Gr'$---if the independence constraints in the two DAGs are
identical. Because equivalence is reflexive, symmetric, and
transitive, the relation defines a set of equivalence classes over
network structures. We will use $\EClass{\Gr}$ to denote the
equivalence class of DAGs to which $\Gr$ belongs.

An equivalence class of DAGs $\Fqclass$ is an {\em independence
  map} ({\em IMAP}) of another
equivalence class of DAGs $\Eqclass$ if all independence constraints
implied by $\Fqclass$ are also implied by $\Eqclass$. For two DAGs
$\Gr$ and $\Hr$, we use $\Gr \leq \Hr$ to denote that $\EClass{\Hr}$
is an IMAP of $\EClass{\Gr}$; we use $\Gr < \Hr$ when $\Gr \leq \Hr$ and
  $\EClass{\Hr} \not = \EClass{\Gr}$.

As shown by Verma and Pearl (1991), two DAGs are equivalent if and
only if they have the same {\em skeleton} (i.e., the graph resulting
from ignoring the directionality of the edges) and the same {\em
  v-structures} (i.e., pairs of edges $X \rightarrow Y$ and $Y \leftarrow Z$
where $X$ and $Z$ are not adjacent).  As a result, we can use a {\em
  partially directed acyclic graph}---or {\em PDAG} for short---to
represent an equivalence class of DAGs: for a PDAG $\pdag$, the
equivalence class of DAGs is the set that share the skeleton and
v-structures with $\pdag$\footnote{The definitions for the skeleton and set of
  v-structures for a PDAG are the obvious extensions to these
  definitions for DAGs.}. 
\nocite{VP91}

We extend our notation for DAG equivalence and the DAG IMAP relation
to include the more general PDAG structure. In particular, for a PDAG
$\pdag$, we use $\EClass{\pdag}$ to denote the corresponding
equivalence class of DAGs. For any pair of PDAGs $\pdag$ and
$\qdag$---where one or both may be a DAG---we use $\pdag \approx
\qdag$ to denote $\EClass{\qdag} = \EClass{\pdag}$
and we use $\pdag \leq \qdag$ to denote $\EClass{\qdag}$ is an IMAP of
$\EClass{\pdag}$. To avoid confusion, for the remainder of the paper
we will reserve the symbols $\Gr$ and $\Hr$ for DAGs.

For any PDAG $\pdag$ and subset of nodes $\bV$, we use
$\ISG{\pdag}{\bV}$ to denote the subgraph of $\pdag$ induced by $\bV$;
that is, $\ISG{\pdag}{\bV}$ has as nodes the set $\bV$ and has as
edges all those from $\pdag$ that connect nodes in $\bV$. We use
$\NeiAdj{X}{Y}$ to denote, within a PDAG, the set of nodes that are
{\em neighbors} of $X$ (i.e., connected with an undirected edge) and
also adjacent to $Y$ (i.e., without regard to whether the connecting
edge is directed or undirected).

An edge in $\Gr$ is {\em compelled} if it exists in every DAG that is
equivalent to $\Gr$. If an edge in $\gr$ is not compelled, we say that
it is {\em reversible}.  A {\em completed} PDAG ({\em CPDAG}) $\cpdag$
is a PDAG with two additional properties: (1) for every directed edge
in $\cpdag$, the corresponding edge in $\Gr$ is compelled and (2) for
every undirected edge in $\cpdag$ the corresponding edge in $\Gr$ is
reversible. Unlike non-completed PDAGs, the CPDAG representation of an
equivalence class is unique.  We use $\Parset{\pdag}{Y}$ to denote the
{\em parents} of node $Y$ in $\pdag$. An edge $X \rightarrow Y$ is
{\em covered} in a DAG if $X$ and $Y$ have the same parents, with the
exception that $X$ is not a parent of itself.

\subsection{Greedy Equivalence Search}

\begin{algorithm}[th]
\SetKwInOut{Input}{Input}
\SetKwInOut{Output}{Output}
\BlankLine
\hrule
\BlankLine
Algorithm $GES(\Data)$
\BlankLine
\hrule
\BlankLine
\Input{Data $\Data$}
\Output{CPDAG $\cpdag$} 
\BlankLine
\Indp
$\cpdag \longleftarrow$ FES($\Data$)\\
$\cpdag \longleftarrow$ BES($\Data$, $\cpdag$)\\
\Return $\cpdag$\\
\Indm
\BlankLine
\hrule
\caption{Pseudo-code for the GES algorithm.} \label{fig:ges}
\end{algorithm}

The GES algorithm, shown in Figure \ref{fig:ges}, performs a two-phase
greedy search through the space of DAG equivalence classes. GES
represents each search state with a CPDAG, and performs transformation
operators to this representation to traverse between states. Each
operator corresponds to a DAG edge modification, and is scored using a
DAG scoring function that we assume has three properties. First, we
assume the scoring function is {\em score equivalent}, which means
that it assigns the same score to equivalent DAGs. Second, we assume
the scoring function is {\em locally consistent}, which means that,
given enough data, (1) if the current state {\em is not} an IMAP of
$\Gr$, the score prefers edge additions that remove incorrect
independences, and (2) if the current state {\em is} an IMAP of $\Gr$,
the score prefers edge deletions that remove incorrect
dependences. Finally, we assume the scoring function is {\em
  decomposable}, which means we can express it as:
\begin{equation} \label{eq:decompose}
Score(\gr, \Data) = \sum_{i=1}^n Score(X_i, \Parset{\gr}{i})
\end{equation}
Note that the data $\Data$ is implicit in the right-hand side Equation
\ref{eq:decompose}.  Most commonly used scores in the literature have
these properties.  For the remainder of this paper, we assume they
hold for the scoring function we use.

All of the CPDAG operators from GES are scored using differences in
the DAG scoring function, and in the limit of large data, these scores
are positive precisely for those operators that remove incorrect
independences and incorrect dependences.

The first phase of the GES---called {\em forward equivalence search}
or {\em FES}---starts with an empty (i.e., no-edge) CPDAG and greedily
applies {\em GES insert} operators until no operator has a positive
score; these operators correspond precisely to the union of all
single-edge additions to all DAG members of the current
(equivalence-class) state. After FES reaches a local maximum, GES
switches to the second phase---called {\em backward equivalence
  search} or {\em BES}---and greedily applies {\em GES delete}
operators until no operator has a positive score; these operators
correspond precisely to the union of all single-edge deletions from
all DAG members of the current state.

\begin{theorem} {\bf (Chickering, 2002)} \label{th:ges}
Let $\cpdag$ be the CPDAG that results from applying
the GES algorithm to $m$ records sampled from a distribution that is
perfect with respect to DAG $\Gr$. Then in the limit of large $m$,
$\cpdag \approx \Gr$.
\end{theorem}

The role of FES in the large-sample limit is only to identify a state
$\cpdag$ for which $\gr \leq \cpdag$; Theorem \ref{th:ges} holds for
GES under any implementation of FES that results in an IMAP of
$\Gr$. The implementation details can be important in practice because
what constitutes a ``large'' amount of data depends on the number of
parameters in the model. In theory, however, we could simply replace
FES with a (constant-time) algorithm that sets $\cpdag$ to be the
no-independence equivalence class.

The focus of our analysis in the next section is on a modified version
of BES, and the details of the delete operator used in this phase are
important. We detail the preconditions, scoring function, and
transformation algorithm for a delete operator in
Figure~\ref{fig:op-delete}. We note that we do not need to make any
CPDAG transformations when {\em scoring} the operators; it is only
once we have identified the highest-scoring (non-negative) delete that
we need to make the transformation shown in the figure. After applying
the edge modifications described in the {\bf foreach} loop, the
resulting PDAG $\pdag$ is not necessarily completed and hence we may
have to convert $\pdag$ into the corresponding CPDAG
representation. As shown by Chickering (2002), this conversion can be
accomplished easily by using the structure of $\pdag$ to extract a DAG
that we then convert into a CPDAG by undirecting all reversible
edges. The complexity of this procedure for a $\pdag$ with $n$ nodes
and $e$ edges is $O(n \cdot e)$, and requires no calls to the scoring
function.

\begin{algorithm}[t]
\SetAlgoNoLine
\SetAlgorithmName{Figure}{figure}{List of Operators}
\SetKwInOut{Preconditions}{$\bullet$ Preconditions}
\SetKwInOut{Transformation}{$\bullet$ Transformation Algorithm} 
\SetKwInOut{ScoreChange}{$\bullet$ Score Change} 
\SetKwInOut{Operator}{Operator} 
\BlankLine
\hrule
\BlankLine
{\bf Operator:}{\hspace{.1in} $Delete(X,Y,\HSet)$ applied to $\cpdag$}
\BlankLine
\hrule
\BlankLine
{\bf $\bullet$ Preconditions} \\
\Indp
\BlankLine
$X$ and $Y$ are adjacent \\ 
$\HSet \subseteq \NeiAdj{Y}{X}$ \\
$\overline{\HSet} = \NeiAdj{Y}{X} \setminus \HSet$ is a clique \\
\Indm
\BlankLine
{\bf $\bullet$ Scoring} \\
\Indp
$Score(Y, \{\Parset{\cpdag}{Y} \cup \overline{\HSet} \} \setminus X) -
Score(Y, X \cup \Parset{\cpdag}{Y} \cup \overline{\HSet})$\\
\Indm
\BlankLine
{\bf $\bullet$ Transformation} \\
\Indp
\BlankLine
Remove edge between $X$ and $Y$\\
\ForEach{$H \in \HSet$}{Replace $Y - H$ with $Y \rightarrow H$\\
\lIf{$X-H$}{Replace with $X \rightarrow H$}
}
Convert to CPDAG\\
\BlankLine
\hrule
\BlankLine
\caption{Preconditions, scoring, and transformation algorithm for a
  delete operator applied to a CPDAG.}
\label{fig:op-delete}
\end{algorithm}

\section{SELECTIVE GREEDY EQUIVALENCE SEARCH} \label{sec:main}

In this section, we define a variant of the GES algorithm called {\em
  selective GES}---or {\em SGES} for short---that uses a subset of the
GES operators. The subset is chosen based on a given property $\Pi$
that is known to hold for the generative structure $\Gr$.  Just like
GES, SGES---shown in Figure~\ref{alg:limited-GES}---has a forward phase
and a backward phase.

For the forward phase of SGES, it suffices for our theoretical
analysis that we use a method that returns an IMAP of $\Gr$ (in the
large-sample limit) using only a polynomial number of insert-operator
score calls. For this reason, we call this phase {\em poly-FES}. A
simple implementation of poly-FES is to return the no-independence
CPDAG (with no score calls), but other implementations are likely more
useful in practice.

The backward phase of SGES---which we call
{\em selective backward equivalence search} ({\em SBES})---uses only a
subset of the BES delete operators. This subset must necessarily
include all {\em $\Pi$-consistent} delete operators---defined
below---in order to maintain the large-sample consistency of GES, but
the subset can (and will) include additional operators for the sake of
efficient enumeration.

The DAG properties used by SGES must be {\em equivalence invariant},
meaning that for any pair of equivalent DAGs, either the property
holds for both of them or it holds for neither of them. Thus, for any
equivalence-invariant DAG property $\Pi$, it makes sense to say that
$\Pi$ either holds or does not hold for a PDAG. As shown by Chickering
(1995), a DAG property is equivalence invariant if and only if it is
invariant to covered-edge reversals; it follows that the property that
each node has at most $k$ parents is equivalence invariant, whereas
the property that the length of the longest directed path is at least
$k$ is not.  Furthermore, the properties for SGES must also be {\em
  hereditary}, which means that if $\Pi$ holds for a PDAG $\pdag$ it
must also hold for all induced subgraphs of $\pdag$. For example, the
max-parent property is hereditary, whereas the property that each node
has {\em at least} $k$ parents is not. We use {\em EIH property} to
refer to a property that is equivalence invariant and hereditary.

\begin{definition} {\noindent \bf $\Pi$-Consistent GES Delete}\\ 
A GES delete operator $Delete(X,Y,\HSet)$ is {\em $\Pi$ consistent}
for CPDAG $\cpdag$ if, for the set of common descendants $\bW$ of $X$
and $Y$ in the resulting CPDAG $\cpdag'$, the property holds for the
induced subgraph $\ISG{\cpdag'}{X \cup Y \cup \bW}$.
\label{def:consistent}
\end{definition}

In other words, after the delete, the property holds for the subgraph
defined by $X$, $Y$, and their common descendants.

\nocite{Chickering95uai}

\begin{algorithm}[h]
\SetKwInOut{Input}{Input}
\SetKwInOut{Output}{Output}
\BlankLine
\hrule
\BlankLine
Algorithm $SGES(\Data, \Pi)$
\BlankLine
\hrule
\BlankLine
\Input{Data $\Data$, Property $\Pi$}
\Output{CPDAG $\cpdag$} 
\BlankLine
\Indp
$\cpdag \longleftarrow$ poly-FES\\
$\cpdag \longleftarrow$ SBES($\Data$, $\cpdag$, $\Pi$)\\
\Return $\cpdag$\\
\Indm
\BlankLine
\hrule
\caption{Pseudo-code for the SGES algorithm.} \label{alg:limited-GES}
\end{algorithm}

\begin{algorithm}[h]
\SetKwInOut{Input}{Input}
\SetKwInOut{Output}{Output}
\SetKwRepeat{ERepeat}{Loop}{End Loop}
\DontPrintSemicolon
\BlankLine
\hrule
\BlankLine
Algorithm $SBES(\Data, \cpdag, \Pi)$
\BlankLine
\hrule
\BlankLine
\Input{Data $\Data$, CPDAG $\cpdag$, Property $\Pi$}
\Output{ CPDAG  } 
\BlankLine
{\bf Repeat} \\
\Indp
${\bf Ops} \longleftarrow$ Generate $\Pi$-consistent delete operators
for \cpdag\\
$Op \longleftarrow$ highest-scoring operator in ${\bf Ops}$\\
\SetAlgoVlined
\lIf{score of $Op$ is negative}{\Return $\cpdag$}
$\cpdag \longleftarrow$ Apply $Op$ to $\cpdag$\\
\Indm
\BlankLine
\hrule
\caption{Pseudo-code for the SBES algorithm.} \label{alg:limited-BES}
\end{algorithm}
 
\subsection{LARGE-SAMPLE CORRECTNESS}

The following theorem establishes a graph-theoretic justification for
considering only the $\Pi$-consistent deletions at each step of
SBES. 

\begin{theorem} \label{th:main}
If $\Gr < \cpdag$ for CPDAG $\cpdag$ and DAG $\Gr$, then for any EIH
property $\Pi$ that holds on $\Gr$, there exists a $\Pi$-consistent
$Delete(X,Y,\HSet)$ that when applied to $\cpdag$ results in the CPDAG
$\cpdag'$ for which $\Gr \leq \cpdag'$.
\end{theorem}

\ArxivUai{
We postpone the proof of Theorem \ref{th:main} to the appendix.
}{
The proof of Theorem \ref{th:main} can be found in Chickering and Meek
(2015), an expanded version of this paper. \nocite{ChickeringMeek2015arxiv}
}
The result is a
consequence of an explicit characterization of, for a given pair of
DAGs $\Gr$ and $\Hr$ such that $\Gr < \Hr$, an edge in $\Hr$ that we
can either reverse or delete in $\Hr$ such that for the resulting DAG
$\Hr'$, we have $\Gr \leq \Hr'$\footnote{Chickering (2002)
  characterizes the {\em reverse} transformation of
  reversals/additions in $\Gr$, which provides an {\em implicit}
  characterization of reversals/deletions in $\Hr$.}.

\begin{theorem} \label{th:ges-limited}
Let $\cpdag$ be the CPDAG that results from applying the SGES
algorithm to (1) $m$ records sampled from a distribution that is
perfect with respect to DAG $\Gr$ and (2) EIH property $\Pi$ that
holds on $\Gr$. Then in the limit of large $m$, $\cpdag \approx \Gr$.
\end{theorem}
{\noindent \bf Proof:} Because the scoring function is locally
consistent, we know poly-FES must return an IMAP of $\Gr$. Because SBES
includes all the $\Pi$-consistent delete operators, Theorem
\ref{th:main} guarantees that, unless $\cpdag \approx \Gr$, there will
be a positive-scoring operator. \qed

\subsection{COMPLEXITY MEASURES}

In this section, we discuss a number of distributional assumptions
that we can use with Theorem \ref{th:ges-limited} to limit the number
of operators that SGES needs to score. As discussed in Section
\ref{sec:related}, when we assume the generative distribution is
perfect with respect to a DAG $\Gr$, then graph-theoretic assumptions
about $\Gr$ can lead to more efficient training algorithms.  Common
assumptions used include (1) a maximum parent-set size for any node,
(2) a maximum-clique\footnote{We use {\em clique} in a DAG to mean a
  set of nodes in which all pairs are adjacent.} size among any nodes
and (3) a maximum treewidth. Treewidth is important because the
complexity of exact inference is exponential in this measure.

We can associate a property with each of these assumptions that holds
precisely when the DAG $\Gr$ satisfies that assumption. Consider the
constraint that the maximum number of parents for any node in $\Gr$ is
some constant $k$. Then, using ``PS'' to denote parent size, we can
define the property $\Pi_{PS}^k$ to be true precisely for those DAGs
in which each node has at most $k$ parents. Similarly we can define
$\Pi_{CL}^k$ and $\Pi_{TW}^k$ to correspond to maximum-clique size and
maximum treewidth, respectively.

For two properties $\Pi$ and $\Pi'$, we write $\Pi \subseteq \Pi'$ if
for every DAG $\Gr$ for which $\Pi$ holds, $\Pi'$ also holds. In other
words, $\Pi$ is a {\em more constraining} property than is
$\Pi'$. Because the lowest node in any clique has all other nodes in
the clique as parents, it is easy to see that $\Pi_{PS}^k \subseteq
\Pi_{CL}^{k-1}$. Because the treewidth for DAG $\Gr$ is defined to be
the size of the largest clique minus one in a graph whose cliques are
at least as large as those in $\Gr$, we also have $\Pi_{TW}^k
\subseteq \Pi_{CL}^{k-1}$.  Which property to use will typically be a
trade-off between how reasonable the assumption is (i.e, less
constraining properties are more reasonable) and the efficiency of the
resulting algorithm (i.e., more constraining properties lead to faster
algorithms).

We now consider a new complexity measure called {\em v-width}, whose
corresponding property is less constraining than the previous three,
and somewhat remarkably leads to an efficient implementation in
SGES. For a DAG $\Gr$, the v-width is defined to be the maximum of,
over all pairs of non-adjacent nodes $X$ and $Y$, the size of the
largest clique among common children of $X$ and $Y$. In other words,
v-width is similar to the maximum-clique-size bound, except that the
bound only applies to cliques of nodes that are shared children of
some pair of non-adjacent nodes. With this understanding it is easy to
see that, for the property $\Pi_{VW}^k$ corresponding to a bound on
the v-width, we have $\Pi_{CL}^k \subseteq \Pi_{VW}^k$.

\begin{figure}[th]
\centering
\begin{minipage}{0.5\textwidth}
\includegraphics[width=3.0in]{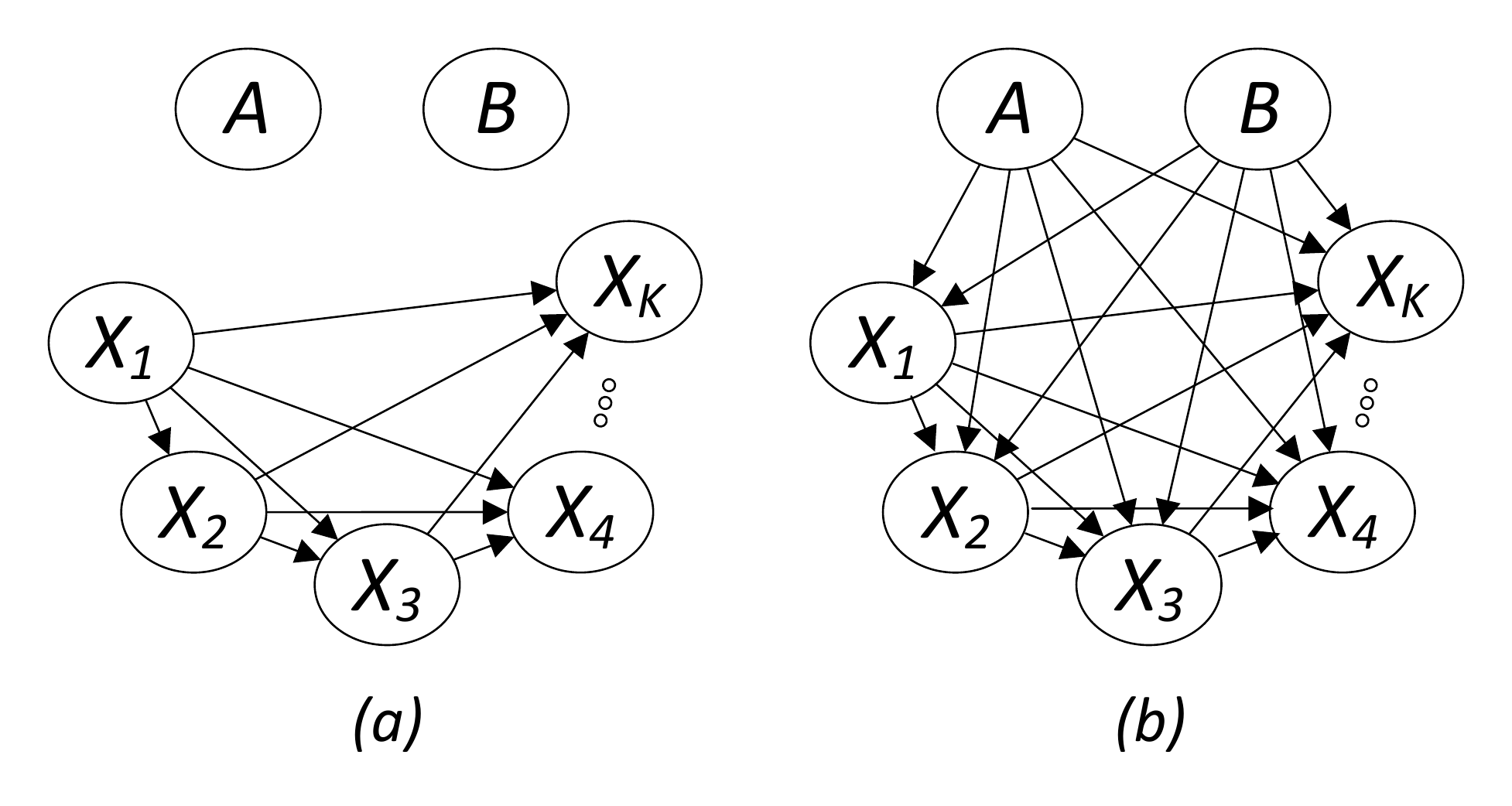}
\end{minipage}
\caption{Two DAGs (a) and (b) having identical maximum clique sizes,
  similar maximum number of parents, and divergent v-widths.}
\label{fig:properties}
\end{figure}

To illustrate the difference between v-width and the other complexity
measures, consider the two DAGs in Figure \ref{fig:properties}. The
DAG in Figure \ref{fig:properties}(a) has a clique of size $K$, and
consequently a maximum-clique size of $K$ and a maximum parent-set
size of $K-1$. Thus, if $K$ is $O(n)$ for a large graph of $n$ nodes,
any algorithm that is exponential in these measures will not be
efficient. The v-width, however, is zero for this DAG. The DAG in
Figure \ref{fig:properties}(b), on the other hand, has a v-width of
$K$.

In order to use a property with SGES, we need to establish that it is
EIH. For $\Pi_{PS}^k$, $\Pi_{CL}^k$ and $\Pi_{VW}^k$,
equivalence-invariance follows from the fact that all three properties
are covered-edge invariant, and hereditary follows because the
corresponding measures cannot increase when we remove nodes and edges
from a DAG. Although we can establish EIH for the treewidth property
$\Pi_{TW}^k$ with more work, we omit further consideration of
treewidth for the sake of space.

\subsection{GENERATING DELETIONS}

In this section, we show how to generate a set of deletion operators
for SBES such that all $\Pi$-consistent deletion operators are
included, for any $\Pi \in \{\Pi_{PS}^k, \Pi_{CL}^k,
\Pi_{VW}^k\}$. Furthermore, the total number of deletion operators we
generate is polynomial in the number of nodes in the domain and
exponential in $k$.

Our approach is to restrict the $Delete(X,Y,\HSet)$ operators based on
the $\HSet$ sets and the resulting CPDAG $\cpdag'$. In particular, we
{\em rule out} candidate $\HSet$ sets for which $\Pi$ does not hold on
the induced subgraph $\ISG{\cpdag'}{\HSet \cup X \cup Y}$; because all
nodes in $\HSet$ will be common children of $X$ and $Y$ in
$\cpdag'$---and thus a subset of the common descendants of $X$ and
$Y$---we know from Definition \ref{def:consistent} (and the fact that
$\Pi$ is hereditary) that none of the dropped operators can be
$\Pi$-consistent.

Before presenting our restricted-enumeration algorithm, we now discuss
how to enumerate delete operators without restrictions.  As shown by
Andersson et al. (1997), a CPDAG is a chain graph whose undirected
components are chordal. This means that the induced sub-graph defined
over $\NeiAdj{Y}{X}$---which is a subset of the neighbors of $Y$---is
an undirected chordal graph. A useful property of chordal graphs is
that we can identify, in polynomial time, a set of {\em maximal
  cliques} over these nodes\footnote{Blair and Peyton (1993) provide a
  good survey on chordal graphs and detail how to identify the maximal
  cliques while running maximum-cardinality search.}; let $\bC_1, ...,
\bC_m$ denote the nodes contained within these $m$ maximal cliques,
and let $\overline{\HSet} = \NeiAdj{Y}{X} \setminus \HSet$ be the
complement of the shared neighbors with respect to the candidate
$\HSet$.  Recall from Figure \ref{fig:op-delete} that the
preconditions for any $Delete(X,Y,\HSet)$ include the requirement that
$\overline{\HSet}$ is a clique. This means that for any valid $\HSet$,
there must be some maximal clique $\bC_i$ that contains the entirety
of $\overline{\HSet}$; thus, we can generate all operators (without
regard to any property) by stepping through each maximal clique
$\bC_i$ in turn, initializing $\HSet$ to be all nodes {\em not} in
$\bC_i$, and then generating a new operator corresponding to expanding
$\HSet$ by all subsets of nodes in $\bC_i$. Note that if
$\NeiAdj{Y}{X}$ is itself a clique, we are enumerating over all
$2^{|\NeiAdj{Y}{X}|}$ operators.

\nocite{BlairPeyton93}
\nocite{AMP97}

As we show below, all three of the properties of interest impose a
bound on the maximum clique size among nodes in $\HSet$. If we
are given such a bound $s$, we know that any ``expansion'' subset for
a clique that has size greater than $s$ will result in an operator
that is not valid. Thus, we can implement the above
operator-enumeration approach more efficiently by only generating
subsets within each clique that have size at most $s$. This allows us
to process each clique $\bC_i$ with only $O(|\bC_i+1|^s)$ calls to the
scoring function. In addition, we need not enumerate over {\em any} of
the subsets of $\bC_i$ if, after removing this clique from the graph,
there remains a clique of size greater than $s$; we define the
function $FilterCliques(\{\bC_1, \ldots, \bC_m\}, s)$ to be the subset
of cliques that remain after imposing this constraint. With this
function, we can define {\sc Selective-Generate-Ops} as shown in
Figure \ref{fig:generate} to leverage the max-clique-size constraint
when generating operators; this algorithm will in turn be used to
generate all of the CPDAG operators during SBES.

\begin{figure}[th]
\centering
\begin{minipage}{0.5\textwidth}
\includegraphics[width=3.0in]{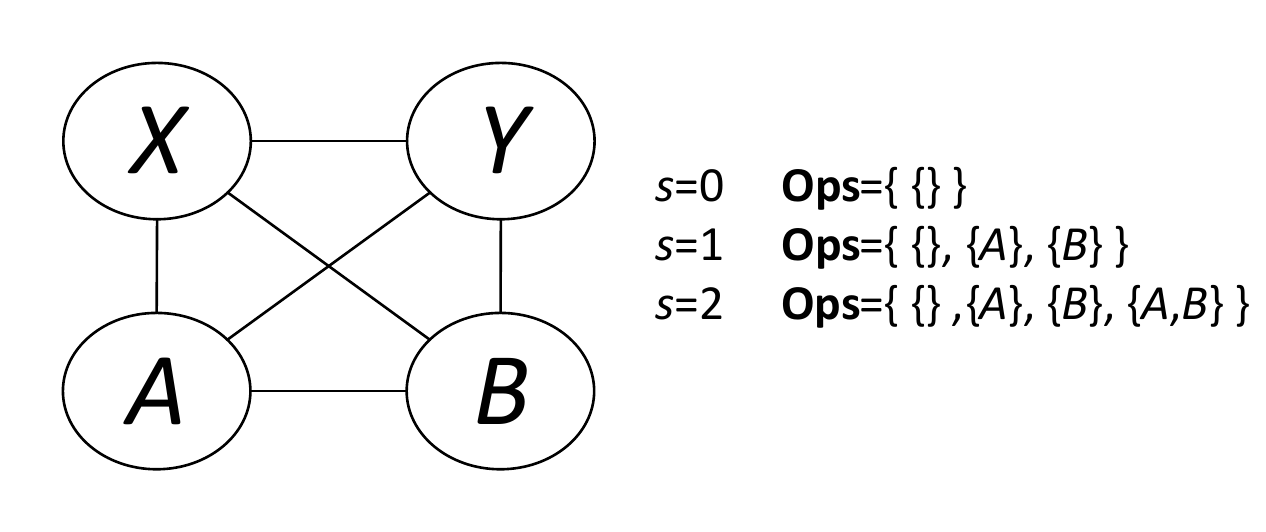}
\end{minipage}
\caption{An example CPDAG $\cpdag$ and the resulting operators
  generated by {\sc Selective-Generate-Ops}($\cpdag$,$X$,$Y$,$s$)
  for various values of $s$.}
\label{fig:GenerateExample}
\end{figure}

{\noindent \bf Example:} In Figure~\ref{fig:GenerateExample}, we show
an example CPDAG for which to run {\sc
  Selective-Generate-Ops}($\cpdag$, $X$, $Y$, $s$) for various values
of $s$. In the example, there is a single clique $\bC = \{A, B\}$ in
the set $\NeiAdj{Y}{X}$, and thus at the top of the outer {\bf foreach}
loop, the set $\HSet_0$ is initialized to the empty set. If $s=0$, the
only subset of $\bC$ with size zero is the empty set, and so that is
added to ${\bf Ops}$ and the algorithm returns. If $s=1$ we add, in
addition to the empty set, all singleton subsets of $\bC$. For $s \geq
2$, we add all subsets of $\bC$. \qed
 
Now we discuss how each of the three properties impose a constraint
$s$ on the maximum clique among nodes in $\HSet$, and consequently the
selective-generation algorithm in Figure \ref{fig:generate} can be
used with each one, given an appropriate bound $s$.  For both
$\Pi_{VW}^k$ and $\Pi_{CL}^k$, the $k$ given imposes an explicit bound
on $s$ (i.e., $s = k$ for both). Because any clique in $\HSet$ of size
$r$ will result in a DAG member of the resulting equivalence class
having a node in that clique with at least $r+1$ parents (i.e., $r-1$
from the other nodes in the clique, plus both $X$ and $Y$), we have
for $\Pi_{PS}^k$, $s = k-1$.

\begin{algorithm}[t]
\SetKwInOut{Input}{Input}
\SetKwInOut{Output}{Output}
\BlankLine
\hrule
\BlankLine
Algorithm {\sc Selective-Generate-Ops}($\cpdag, X, Y, s$)
\BlankLine
\hrule
\BlankLine\Input{CPDAG $\cpdag$ with adjacent $X$,$Y$ and limit $s$}
\Output{${\bf Ops} = \{\HSet_{1}, \ldots, \HSet_{m}$\}} 
\Indp
\BlankLine 
${\bf Ops} \longleftarrow \emptyset$\\
Generate maximal cliques  $\bC_1, ..., \bC_m$ from $\NeiAdj{Y}{X}$\\
$\bS \longleftarrow FiltertCliques(\{\bC_1, \ldots, \bC_m\}, s)$\\
\ForEach{$\bC_i \in \bS$}
        {
          $\HSet_0 \longleftarrow \NeiAdj{Y}{X} \setminus \bC_i$\\
          \ForEach{$\bC \subseteq \bC_i$ with $|\bC| \leq s$}
                  {
                     Add $\HSet_0 \cup \bC$ to ${\bf Ops}$
                  }
        }
\Return{${\bf Ops}$}\\
\BlankLine
\hrule
\BlankLine
\Indm
\caption{Algorithm to generate clique-size limited delete operators.}
\label{fig:generate}
\end{algorithm}

We summarize the discussion above in the following proposition.

\begin{proposition}
Algorithm {\sc Selective-Generate-Ops} applied to all edges using
clique-size bound $s$ generates all $\Pi$-consistent delete operators
for $\Pi \in \{\Pi_{PS}^{s+1}, \Pi_{CL}^s, \Pi_{VW}^s\}$.
\end{proposition}

We now argue that running SBES on a domain of $n$ variables when using
Algorithm {\sc Selective-Generate-Ops} with a bound $s$ requires only a
polynomial number in $n$ of calls to the scoring function. Each clique
in the inner loop of the algorithm can contain at most $n$ nodes, and
therefore we generate and score at most $(n+1)^s$ operators, requiring
at most $2(n+1)^s$ calls to the scoring function. Because the cliques
are maximal, there can be at most $n$ of them considered in the outer
loop. Because there are never more than $n^2$ edges in a CPDAG, and we
will delete at most all of them, we conclude that even if we decided
to rescore every operator after every edge deletion, we will only make
a polynomial number of calls to the scoring function.

From the above discussion and the fact that SBES completes using at
most a polynomial number of calls to the scoring function, we get the
following result for the full SGES algorithm.

\begin{proposition} \label{prop:complex-complete}
The SGES algorithm, when run over a domain of $n$ variables and given
$\Pi \in \{\Pi_{PS}^{s+1}, \Pi_{CL}^s, \Pi_{VW}^s\}$, runs to
completion using a number of calls to the DAG scoring function that is
polynomial in $n$ and exponential in $s$.
\end{proposition}

\section{EXPERIMENTS} \label{sec:experiments}

In this section, we present a simple synthetic experiment comparing
SBES and BES that demonstrates the value of pruning operators.  In our
experiment we used an {\em oracle} scoring function. In particular,
given a generative model $\gr$, our scoring function computes the
minimum-description-length score assuming a data size of five billion
records, but without actually sampling any data: instead, we use exact
inference in $\gr$ (i.e., instead of counting from data) to compute
the conditional probabilities needed to compute the expected log
loss. This allows us to get near-asymptotic behavior without the need
to sample data. To evaluate the cost of running each algorithm, we
counted the number of times the scoring function was called on a
unique node and parent-set combination; we cached these scores away so
that if they were needed multiple times during a run of the algorithm,
they were only computed (and counted) once.

\begin{figure}[th]
\centering
\begin{minipage}{0.5\textwidth}
\includegraphics[width=3.0in]{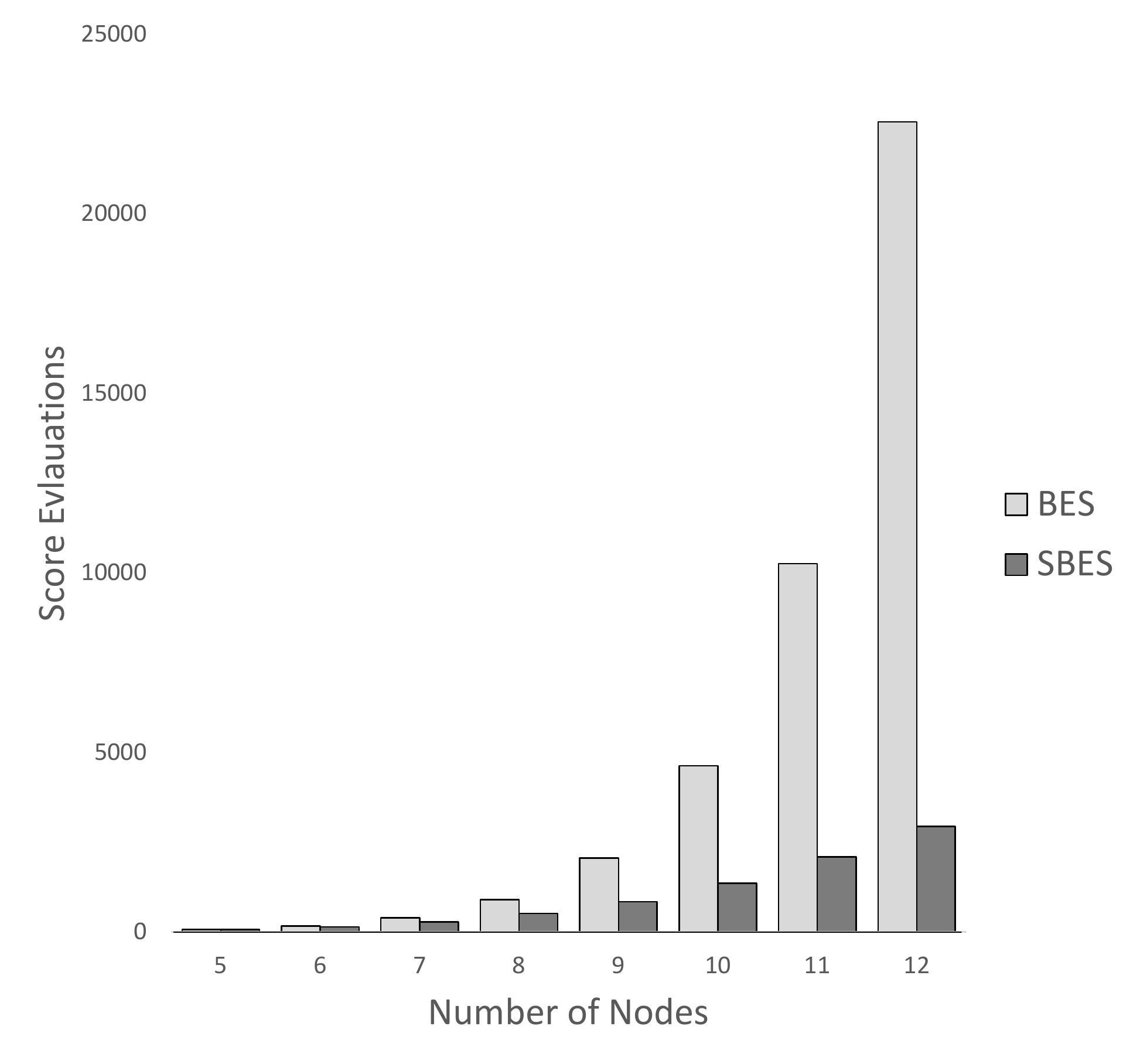}
\end{minipage}
\caption{Number of score evaluations needed to run BES and SBES,
  starting from the complete graph, for a range of domain sizes.}
\label{fig:clique}
\end{figure}

In Figure \ref{fig:clique}, we show the average number of
scoring-function calls required to complete BES and SBES when starting
from a complete graph over a domain of $n$ binary variables, for
varying values of $n$. Each average is taken over ten trials,
corresponding to ten random generative models. All variables in the
domain were binary.  We generated the structure of each generative
model as follows. First, we enumerated all node pairs by randomly
permuting the nodes and taking each node in turn with all of its
predecessors in turn. For each node pair in turn, we chose to attempt
an edge insertion with probability one half. For each attempt, we
added an edge if doing so (1) did not create a cycle and (2) did not
result in a node having more than two parents; if an edge could be
added in either direction, we chose the direction at random. We
sampled the conditional distributions for each node and each parent
configuration from a uniform Dirichlet distribution with
equivalent-sample size of one. We ran SBES with $\Pi_{PS}^2$.

Our results show clearly the exponential dependence of BES on the
number of nodes in the clique, and the increasing savings we get with
SBES, leveraging the fact that $\Pi_{PS}^2$ holds in the generative
structure.

Note that to realize large savings in practice, when GES runs FES
instead of starting from a dense graph, a (relatively sparse)
generative distribution must lead FES to an equivalence class
containing a (relatively dense) undirected clique that is subsequently
``thinned'' during BES. We can synthesize challenging grid
distributions to force FES into such states, but it is not clear how
realistic such distributions are in practice. When we re-run the
clique experiment above, but where we instead start both BES and SBES
from the model that results from running FES (i.e., with no
polynomial-time guarantee), the savings from SBES are small due to
the fact that the subsequent equivalence classes do not contain large
cliques.

\section{CONCLUSION} \label{sec:conclude}

Through our selective greedy equivalence search algorithm SGES, we
have demonstrated how to leverage graph-theoretic properties to reduce
the need to score graphs during score-based search over equivalence
classes of Bayesian networks. Furthermore, we have shown that for
graph-theoretic complexity properties including maximum-clique size,
maximum number of parents, and v-width, we can guarantee that the number
of score evaluations is polynomial in the number of nodes and
exponential in these complexity measures.

The fact that we can use our approach to selectively choose operators for any 
hereditary and equivalence invariant graph-theoretic property provides the 
opportunity to explore alternative complexity measures. Another 
candidate complexity measure is the maximum number of
v-structures. Although the corresponding property does not limit the
maximum size of a {\em clique} in $\HSet$, it limits directly the size
$|\HSet|$ for every operator. Thus it would be easy to enumerate
these operators efficiently. Another complexity measure of interest is
{\em treewidth}, due to the fact that exact inference in a
Bayesian-network model is takes time exponential in this measure. 

The results we have presented are for the general Bayesian-network
learning problem.  It is interesting to consider the implications of
our results for the problem of learning particular subsets of Bayesian
networks. One natural class that we discussed in
Section~\ref{sec:related} is that of polytrees. If we assume that the
generative distribution is perfect with respect to a polytree then we
know the v-width of the generative graph is one. This implies, in the
limit of large data, that we can recover the structure of the
generative graph with a polynomial number of score evaluations. This
provides a score-based recovery algorithm analogous to the
constraint-based approach of Geiger et al. (1990).

We presented a simple complexity analysis for the purpose of
demonstrating that SGES uses a only polynomial number of calls to the
scoring function. We leave as future work a more careful analysis that
establishes useful constants in this polynomial. In particular, we can
derive tighter bounds on the total number of
node-and-parent-configurations that are needed to score all the
operators for each CPDAG, and by caching these configuration scores we
can further take advantage of the fact that most operators remain
valid (i.e., the preconditions still hold) and have the same score
after each transformation.

Finally, we plan to investigate practical implementations of
poly-FES that have the polynomial-time guarantees needed for SGES.

\appendix
\section*{Appendices}

In the following two appendices, we prove Theorem \ref{th:main}.

\section{Additional Background}

In this section, we introduce additional background material needed
for the proofs.

\subsection{Additional Notation} \label{sec:more-notation}

To express sets of variables more compactly, we often use a comma to
denote set union (e.g., we write $\bX = \bY, \bZ$ as a more compact
version of $\bX = \bY \cup \bZ$). We also will sometimes remove the
comma (e.g., $\bY\bZ$).  When a set consists of a singleton variable,
we often use the variable name as shorthand for the set containing
that variable (e.g., we write $\bX = \bY \setminus Z$ as shorthand for
$\bX = \bY \setminus \{Z\}$).

We say a node $N$ is a {\em descendant} of $Y$ if $N = Y$ or there is
a directed path from $Y$ to $N$. We use $\Hr$-descendant to refer to a
descendant in a particular DAG $\Hr$. We say a node $N$ is a {\em
  proper descendant} of $Y$ if $N$ is a descendant of $Y$ and $N \not
= Y$. We use $\NonDesc{\Hr}{Y}$ to denote the non-descendants of node
$Y$ in $\Gr$.  We use $\Parnot{Y}{X_1X_2 \ldots X_n}$ as shorthand for
$\Parset{\Hr}{Y} \setminus \{X_1 , \ldots , X_n\}$. For example, to
denote all the parents of $Z$ in $\Hr$ except for $X$ and $Y$, we use
$\Parnot{Z}{XY}$.

\subsection{D-separation and Acvite Paths}

The independence constraints implied by a DAG structure are
characterized by the {\em d-separation} criterion. Two nodes $A$ and
$B$ are said to be d-separated in a DAG $\gr$ given a set of nodes
$\SSet$ if and only if there is no {\em active path} in $\gr$ between
$A$ and $B$ given $\SSet$. The standard definition of an active path
is a {\em simple} path for which each node $W$ along the path either
(1) has converging arrows (i.e., $\rightarrow W \leftarrow$) and $W$
or a descendant of $W$ is in $\SSet$ or (2) does not have converging
arrows and $W$ is not in $\SSet$. By simple, we mean that the path
never passes through the same node twice.

To simplify our proofs, we use an equivalent definition of an active
path---that need not be simple---where each node $W$ along the path
either (1) has converging arrows and $W$ is in $\SSet$ or (2) does not
have converging arrows and $W$ is not in $\SSet$. In other words,
instead of allowing a segment $\rightarrow W \leftarrow$ to be
included in a path by virtue of a descendant of $W$ belonging to
$\SSet$, we require that the path include the sequence of edges from
$W$ to that descendant and then back again. For those readers familiar
with the celebrated ``Bayes ball'' algorithm of Shachter (1998) for
testing d-separation, our expanded definition of an active path is
simply a valid path that the ball can take between $A$ and
$B$. \nocite{Shachter98}

We use $\bX \indep_{\Gr} \bY | \bZ$ to denote the assertion that DAG
$\gr$ imposes the constraint that variables $\bX$ are independent of
variables $\bY$ given variables $\bZ$.When a node $W$ along a path has
converging arrows, we say that $W$ is a {\em collider} at that
position in the path.

The direction of each {\em terminal} edge in an active path---that
is, the first and last edge encountered in a traversal from one end of
the path to the other---is important for determining whether we can
append two active paths together to make a third active path. We say
that a path $\pi(A,B)$ is {\em into} $A$ if the terminal edge incident
to $A$ is oriented toward $A$ (i.e., $A \leftarrow$). Similarly, the
path is into $B$ if the terminal edge incident to $B$ is oriented
toward $B$. If a path is not into an endpoint $A$, we say that the
path is {\em out of} $A$. Using the following result from Chickering
(2002), we can combine active paths together.
\nocite{Chickering02JMLRb}

\begin{lemma} {\bf (Chickering, 2002)} \label{lem:append}
Let $\pi(A,B)$ be an $\SSet$-active path between $A$ and $B$, and let
$\pi(B,C)$ be an $\SSet$-active path between $B$ and $C$. If either
path is out of $B$, then the concatenation of $\pi(A,B)$ and
$\pi(B,C)$ is an $\SSet$-active path between $A$ and $C$.
\end{lemma}
\nocite{Chickering02JMLRb}

Given a DAG $\Hr$ that is an IMAP of DAG $\Gr$, we use the
d-separation criterion in two general ways in our proofs. First, we
identify d-separation facts that hold in $\Hr$ and conclude that they
must also hold in $\Gr$. Second, we identify active paths in $\Gr$ and
conclude that there must be corresponding active paths in $\Hr$.

\subsection{Independence Axioms}

In many of our proofs, we would like to reason about the independence
facts that hold in DAG $\gr$ without knowing what its structure is,
which makes using the d-separation criterion problematic. As described
in Pearl (1988), any set of independence facts characterized by the
d-separation criterion also respect the independence axioms shown in
Figure \ref{fig:axioms}. These axioms allow us to take a set of
independence facts in some unknown $\Gr$ (e.g., that are implied by
d-separation in $\Hr$), and derive new independence facts that we
know must also hold in $\Gr$.
\nocite{Pearl88}

\comment{
\begin{figure*}[th]
\begin{tabular}[l]{lrcl}
\hspace{.01in} \\
\multicolumn{2}{l}{\em Symmetry:} \\
& \cind{\bX}{\bY}{\bZ} & $\Longleftrightarrow$ & \cind{\bY}{\bX}{\bZ} \\
\multicolumn{2}{l}{\em Decomposition:} \\
& \cind{\bX}{\bY,\bW}{\bZ} & $\Longrightarrow$ & \cind{\bX}{\bY}{\bZ} \axand \cind{\bX}{\bW}{\bZ}  \\
\multicolumn{2}{l}{\em Composition:} \\
& \cind{\bX}{\bY,\bW}{\bZ} & $\Longleftarrow$ & \cind{\bX}{\bY}{\bZ}
\axand \cind{\bX}{\bW}{\bZ}  \\
\multicolumn{2}{l}{\em Intersection:} \\
& \cind{\bX}{\bY,\bW}{\bZ}  
& $\Longleftarrow$ 
& \cind{\bX}{\bY}{\bZ,\bW} \axand \cind{\bX}{\bW}{\bZ,\bY} 
\\
\multicolumn{2}{l}{\em Weak Union:} \\
& \cind{\bX}{\bY,\bW}{\bZ} & $\Longrightarrow$ & \cind{\bX}{\bY}{\bZ,\bW}  \\
\multicolumn{2}{l}{\em Contraction:} \\
& \cind{\bX}{\bY,\bW}{\bZ}  
& $\Longleftarrow$ 
& \cind{\bX}{\bW}{\bZ,\bY} \axand \cind{\bX}{\bY}{\bZ} 
\\
\multicolumn{2}{l}{\em Weak Transitivity:} \\[.1in]
& \cind{\bX}{\bY}{\bZ} \axand \cind{\bX}{\bY}{\bZ,T} &
$\Longrightarrow$ & \cind{\bX}{T}{\bZ} \axor \cind{\bY}{T}{\bZ}  \\
\end{tabular}
\caption{Independence Axioms}
\end{figure*}
}

\begin{figure*}[th]
\begin{tabular}[l]{llrcl}
\hspace{.01in} \\
\multicolumn{2}{l}{\em Symmetry:} 
& \cind{\bX}{\bY}{\bZ} & $\Longleftrightarrow$ & \cind{\bY}{\bX}{\bZ} \\
\multicolumn{2}{l}{\em Decomposition:} 
& \cind{\bX}{\bY,\bW}{\bZ} & $\Longrightarrow$ & \cind{\bX}{\bY}{\bZ} \axand \cind{\bX}{\bW}{\bZ}  \\
\multicolumn{2}{l}{\em Composition:} 
 & \cind{\bX}{\bY}{\bZ}
\axand \cind{\bX}{\bW}{\bZ}  
 & $\Longrightarrow$
& \cind{\bX}{\bY,\bW}{\bZ}
\\
\multicolumn{2}{l}{\em Intersection:} 
& \cind{\bX}{\bY}{\bZ,\bW} \axand \cind{\bX}{\bW}{\bZ,\bY} 
& $\Longrightarrow$ 
& \cind{\bX}{\bY,\bW}{\bZ}  
\\
\multicolumn{2}{l}{\em Weak Union:} 
& \cind{\bX}{\bY,\bW}{\bZ} & $\Longrightarrow$ & \cind{\bX}{\bY}{\bZ,\bW}  \\
\multicolumn{2}{l}{\em Contraction:} 
& \cind{\bX}{\bW}{\bZ,\bY} \axand \cind{\bX}{\bY}{\bZ} 
& $\Longleftarrow$ 
& \cind{\bX}{\bY,\bW}{\bZ}  
\\
\multicolumn{2}{l}{\em Weak Transitivity:} 
& \cind{\bX}{\bY}{\bZ} \axand \cind{\bX}{\bY}{\bZ,T} &
$\Longrightarrow$ & \cind{\bX}{T}{\bZ} \axor \cind{\bY}{T}{\bZ}  \\
\end{tabular}
\caption{The DAG-perfect independence axioms.}
\label{fig:axioms}
\end{figure*}

Throughout the proofs, we will often use the Symmetry axiom
implicitly. For example, if we have $\cind{A}{B,C}{D}$ we might claim
that $\cind{B}{A}{C,D}$ follows from Weak Union, as opposed to
concluding $\cind{A}{B}{C,D}$ from Weak Union and then applying
Symmetry.  We will frequently identify independence constraints in
$\Hr$ and conclude that they hold in $\Gr$, without explicitly
justifying this with {\em because $\Gr \leq \Hr$}. For example, we
will say: \\

{\em Because $A$ is a non-descendant of $B$ in $\Hr$, it follows from
  the Markov conditions that
  $\gcind{A}{B}{\Parset{\Hr}{B}}$.}\\

In other words, to be explicit we would say that
$\hcind{A}{B}{\Parset{\Hr}{B}}$ follows from the Markov conditions,
and the independence holds in $\Gr$ because $\Gr \leq \Hr$. 

The Composition axiom states that if $\bX$ is independent of both
$\bY$ and $\bW$ individually given $\bZ$, then $\bX$ is independent of
them jointly. If we have more than two such sets that are independent
of $\bX$, we can apply the Composition axiom repeatedly to combine
them all together. To simplify, we will do this combination
implicitly, and assume that the Composition axiom is defined more
generally. Thus, for example, we might have:

{\em Because $\cind{X}{Y}{\bZ}$ for every $Y \in \bY$, we conclude by
  the Composition axiom that $\cind{X}{\bY}{\bZ}$.}

\section{Proofs} \label{sec:proofs}

In this section, we provide a number of intermediate results that lead
to a proof of Theorem \ref{th:main}.

\subsection{Intermediate Result: ``The Deletion Lemma''}

Given DAGs $\gr$ and $\Hr$ for which $\gr < \Hr$, we say that an edge
$e$ from $\Hr$ is {\em deletable in $\Hr$ with respect to $\Gr$} if,
for the DAG $\Hr'$ that results after removing $e$ from $\Hr$, we have
$\Gr \leq \Hr$. We will say that an edge is {\em deletable in $\Hr$}
or simply {\em deletable} if $\Gr$ or both DAGs, respectively, are
clear from context. The following lemma establishes necessary and
sufficient conditions for an edge to be deletable. 

\begin{figure*}[th]
\centering
\begin{minipage}{1.0\textwidth}
\includegraphics[width=6.0in]{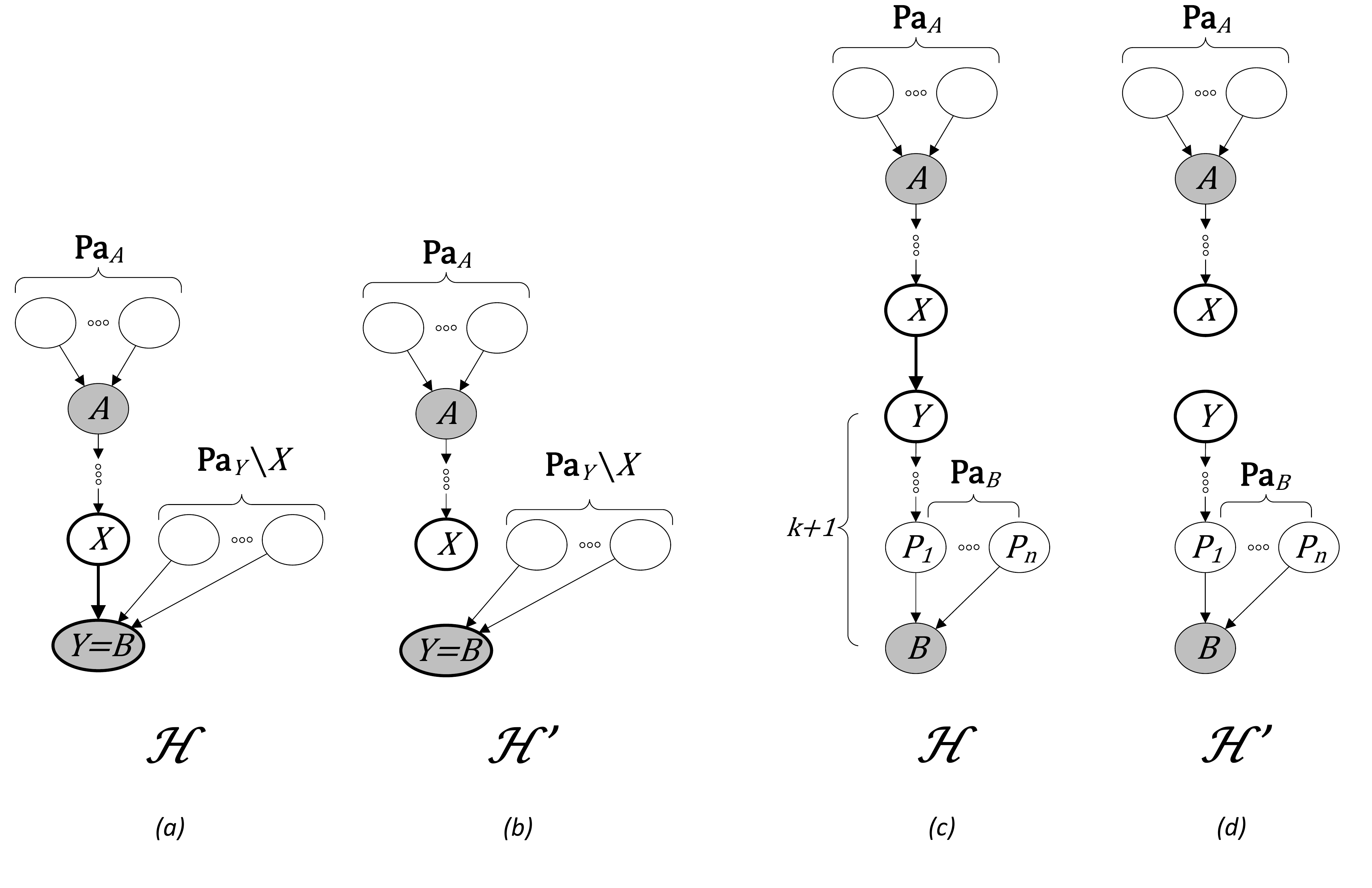}
\end{minipage}
\caption{Relevant portions of $\Hr$ and $\Hr'$ for the inductive proof
  of Lemma \ref{lem:delete}: (a) and (b) are for the basis and (c) and (d) are
  for the induction hypothesis.}
\label{fig:DeletableProof}
\end{figure*}

\begin{lemma} \label{lem:delete}
Let $\Gr$ and $\Hr$ be two DAGs such that $\Gr \leq \Hr$. An edge $X
\rightarrow Y$ is deletable in $\Hr$ with respect to $\Gr$ if and only
if $Y \indep_{\Gr} X | \Parset{\Hr}{Y} \setminus X$.
\end{lemma}
{\noindent \bf Proof:} Let $\Hr'$ be the DAG resulting from removing
the edge. The ``only if'' follows immediately because the given
independence is implied by $\Hr'$.  For the ``if'', we show that for
every node $A$ and every node $B \in \NonDesc{\Hr'}{A}$, the independence
$A \indep_{\gr} B | \Parset{\Hr'}{A}$ holds (in $\gr$). We need only
consider $(A,B)$ pairs for which $B$ is a descendant in $\Hr$ but not
in $\Hr'$; if the ``descendant'' relationship has not changed, we know
the independence holds by virtue of $\gr \leq \Hr$ and the fact that
deleting an edge results in strictly more independence constraints.

The proof follows by induction on the length of the longest directed
path in $\Hr'$ from $Y$ to $B$. For the base case (see Figure
\ref{fig:DeletableProof}a and Figure \ref{fig:DeletableProof}b), we
start with a longest path of length zero; in other words, $B =
Y$. Because $A$ is an ancestor of $Y$ in $\Hr$, both it and its
parents must be non-descendants of $Y$ in $\Hr$, and therefore the
Markov conditions in $\Hr$ imply
\begin{equation} \label{eq:del1}
Y \indep_{\gr} A, \Parset{\Hr}{A} | \Parset{\Hr}{Y}
\end{equation}
Given the independence fact assumed in the lemma, we can apply the
Contraction axiom to remove $X$ from the conditioning set in
(\ref{eq:del1}), and then apply the Weak Union axiom to move
$\Parset{\Hr}{A}$ into the conditioning set to conclude
\begin{equation} \label{eq:del2}
Y \indep_{\gr} A | \Parset{\Hr}{Y} \setminus X, \Parset{\Hr}{A} 
\end{equation}
Neither $Y$ nor its new parents $\Parset{\Hr}{Y} \setminus X$ can be
descendants of $A$ in $\Hr'$, else $B$ would remain a descendant of
$A$ after the deletion, and thus we conclude by the Markov conditions
in $\Hr$ that
\begin{equation} \label{eq:del3}
A \indep_{\gr} \Parset{\Hr}{Y} \setminus X | \Parset{\Hr}{A}
\end{equation}
Applying the Contraction axiom to
(\ref{eq:del2}) and (\ref{eq:del3}), we have 
$$
A \indep_{\gr} Y | \Parset{\Hr}{A}
$$
and because $\Parset{\Hr'}{A} = \Parset{\Hr}{A}$ the lemma follows.

For the induction step (see Figure \ref{fig:DeletableProof}c and
Figure \ref{fig:DeletableProof}d), we assume the lemma holds for all
nodes whose longest path from $Y$ is $\leq k$, and we consider a $B$
for which the longest path from $Y$ is $k+1$. Consider any parent $P$
of node $B$. If $P$ is a descendant of $Y$, the longest path from $Y$
to $B$ must be $\leq k$, else we have a path to $B$ that is longer
than $k+1$. If $P$ is not a descendant of $Y$, then $P$ is also not a
descendant of $A$ in $\Hr$, else $B$ would be a descendant of $A$ in
$\Hr'$. Thus, for every parent $P$, we conclude
$$
A \indep_{\gr} P | \Parset{\Hr}{A}
$$ either by the induction hypothesis or by the fact that $P$ is a
non-descendant of $A$ in $\Hr$. From the Composition axiom we can
combine these individual parents together, yielding
\begin{equation} \label{eq:del4}
A \indep_{\gr} \Parset{\Hr}{B} | \Parset{\Hr}{A}
\end{equation}
Because $B$ is a descendant of $A$ in $\Hr$, we know that $A$ and all
of its parents $\Parset{\Hr}{A}$ are non-descendants of $B$ in $\Hr$,
and thus
\begin{equation} \label{eq:del5}
B \indep_{\gr} A, \Parset{\Hr}{A} | \Parset{\Hr}{B}
\end{equation}
Applying the Weak Union axiom to (\ref{eq:del5}) yields
\begin{equation} \label{eq:del6}
B \indep_{\gr} A | \Parset{\Hr}{A}, \Parset{\Hr}{B}
\end{equation}
and finally applying the Contraction axiom to (\ref{eq:del4}) and
(\ref{eq:del6}) yields
$$
A \indep_{\gr} B | \Parset{\Hr}{A}
$$
Because the parents of $A$ are the same in both $\Hr$ and $\Hr'$, the
lemma follows. \qed

\subsection{Intermediate Result: ``The Deletion Theorem''}

We define the {\em pruned variables for $\Gr$ and $\Hr$}---denoted
$\prunegh$-- to be the subset of the variables that remain after we
repeatedly remove from both graphs any common sink nodes (i.e., nodes
with no children) with the same parents in both graphs.  For $\bV =
\prunegh$, let ${\overline{\bV}}$ denote the complement of $\bV$. Note
that every node in ${\overline{\bV}}$ has the same parents and
children in both $\Gr$ and $\Hr$, and that $\ISG{\Gr}{\overline{\bV}}
= \ISG{\Hr}{\overline{\bV}}$.

We use $\gleaf$ to denote any node in $\Gr$ that has no children. For
any $\gleaf$ $L$, we say that $L$ is an {\em $\Hr$-lowest $\gleaf$} if
no proper descendant of $L$ in $\Hr$ is a $\gleaf$. Note that we are
discussing {\em two} DAGs in this case: $L$ is a leaf in $\Gr$, and
out of all nodes that are leaves in $\Gr$, $L$ is the one that is
lowest in the other DAG $\Hr$. To avoid ambiguity, we often prefix
other common graph concepts (e.g., {\em $\Gr$-child} and {\em
  $\Hr$-parent}) to emphasize the specific DAG to which we are
referring.

We need the following result from Chickering (2002).

\begin{lemma} {\bf (Chickering, 2002)} \label{lem:induction}
Let $\Gr$ and $\Hr$ be two DAGs containing a node $Y$ that is a sink
in both DAGs and for which $\Parset{\Gr}{Y} = \Parset{\Hr}{Y}$. Let
$\Gr'$ and $\Hr'$ denote the subgraphs of $\Gr$ and $\Hr$,
respectively, that result by removing node $Y$ and all its in-coming
edges. Then $\Gr \lessequiv \Hr$ if and only if $\Gr' \lessequiv
\Hr'$.
\end{lemma}

By repeatedly applying Lemma \ref{lem:induction}, the following
corollary follows immediately.

\begin{corollary} \label{cor:induction}
Let $\bV = \prunegh$. Then $\Gr \leq \Hr$ if and only if $\Gr_{\bV} \leq
\Hr_{\bV}$.
\end{corollary}

We now present the ``deletion theorem'', which is the basis for
Theorem \ref{th:main}.

\begin{theorem} \label{th:delete}
Let $\Gr$ and $\Hr$ be DAGs for which $\Gr \leq \Hr$, let $\bV =
\prunegh$, and let $L$ be any $\ISG{\Hr}{\bV}$-lowest
$\ISG{\Gr}{\bV}$-leaf. Then,
\begin{enumerate}
\item If $L$ does not have any $\ISG{\Hr}{\bV}$-children, then for every
  $D \in \bV$ that is an $\ISG{\Hr}{\bV}$-parent of $L$ but not a $\ISG{\Gr}{\bV}$-parent of
  $L$, $D \rightarrow L$ is deletable in $\Hr$.
\item If $L$ has at least one $\ISG{\Hr}{\bV}$-child, let $A$ be any
  $\ISG{\Hr}{\bV}$-highest child; one of the following three properties
  must hold in $\Hr$:
\begin{enumerate}
\item $L \rightarrow A$ is covered.
\item There exists an edge $A \leftarrow B$, where $L$ and $B$ are not adjacent, and either $L \rightarrow
  A$ or $A \leftarrow B$ (or both) are deletable.
\item There exists an edge $D \rightarrow L$, where $D$ and $A$ are not adjacent, and either $D \rightarrow
  L$ or $L \rightarrow A$ (or both) are deletable.
\end{enumerate}
\end{enumerate}
\end{theorem}
{\noindent \bf Proof:} As a consequence of Corollary
\ref{cor:induction}, the lemma holds if and only if it holds for any
graphs $\Gr$ and $\Hr$ for which there are no nodes that are sinks in
both graphs with the same parents; in other words, $\Gr = \Gr_{\bV}$
and $\Hr = \Hr_{\bV}$. Thus, to vastly simplify the notation for the
remainder of the proof, we will assume that this is the case, and
therefore $L$ is a leaf node in $\Gr$, $A$ is a highest child of $L$
in $\Hr$, and the restriction of $B$ and $D$ to $\bV$ is vacuous.

For case (1), we know that $\Parset{\Gr}{L} \subseteq
\Parset{\Hr}{L}$, else there would be some edge in $X \rightarrow L$
in $\Gr$ for which $X$ and $L$ are not adjacent in $\Hr$,
contradicting $\Gr \leq \Hr$. Because $L$ is a leaf in $\Gr$,
all non-parents must also be non-descendants, and hence $L
\indep_{\gr} X | \Parset{\Gr}{L}$ for all $X$. It follows that for
every $D \in \{\Parset{\Hr}{L} \setminus \Parset{\Gr}{L}\}$, $D
\rightarrow Y$ is deletable in $\Hr$. There must exist such a $D$,
else $L$ would be in $\bV = \prunegh$. 

For case (2), we now show that at least one of the properties must
hold. Assume that the first property {\em does not} hold, and
demonstrate that one of the other two properties must hold.  If the
first property does not hold then we know that in $\Hr$ either there
exists an edge $A \leftarrow B$ where $B$ is not a parent of $L$, or
there exists an edge $D \rightarrow L$ where $D$ is not a parent of
$A$. Thus the pre-conditions of at least one of the remaining two
properties must hold.

Suppose $\Hr$ contains the edge $A \leftarrow B$ where $B$ is not a
parent of $L$. Then we conclude immediately from Corollary
\ref{cor:yab} that either $L \rightarrow A$ or $A \leftarrow B$ is
deletable in $\Hr$. 

Suppose $\Hr$ contains the edge $D \rightarrow L$ where $D$ is not a
parent of $A$. Then the set $\bD$ containing {\em all} parents of $L$
that are not parents of $A$ is non-empty. Let $\bR = \Parset{\Hr}{A}
\cap \Parset{\Hr}{L}$ be the shared parents of $L$ and $A$, and let
$\bT = \Parnot{A}{\bR L}$ be the remaining non-$L$ parents of $A$ in
$\Hr$, so that we have $ \Parset{\Hr}{A} = L , \bR , \bT $ and
$\Parset{\Hr}{L} = \bR , \bD $.  Because no node in $\bD$ is a
child or a descendant of $A$, lest $\Hr$ contains a cycle, we know
that $\Hr$ contains the following independence constraint that must
hold in $\Gr$:
\begin{equation} \label{eq:dya-with-L}
A \indep_{\Gr} \bD | L , \bR , \bT
\end{equation}
Because $L$ is a leaf node in $\Gr$, it is impossible to create a new
active path by removing it from the conditioning set, and hence we
also know
\begin{equation} \label{eq:dya-no-L}
A \indep_{\Gr} \bD | \bR , \bT
\end{equation}
Applying the Weak Transitivity axiom to Independence
\ref{eq:dya-with-L} and Independence \ref{eq:dya-no-L}, we conclude
either $A \indep_{\Gr} L | \bR , \bT$---in which case $L
\rightarrow A$ is deletable--or
\begin{equation} \label{eq:ad}
L \indep_{\Gr} \bD | \bR , \bT
\end{equation}
We know that no node in $\bT$ can be a descendant of $L$, or else $A$
would not be the highest child of $L$. Thus, because $L$ is
independent of any non-descendants given its parents we have
\begin{equation} \label{eq:lt}
L \indep_{\Gr} \bT | \bR , \bD
\end{equation}
Applying the Intersection axiom to Independence \ref{eq:ad} and
Independence \ref{eq:lt}, we have
\begin{equation}
L \indep_{\gr} \bD | \bR
\end{equation}
In other words, $L$ is independent of {\em all} of the nodes in $\bD$
given the other parents. By applying the Weak Union axiom, we can pull
all but one of the nodes in $\bD$ into the conditioning set to obtain
\begin{equation}
L \indep_{\gr} D | \bR , \{\bD \setminus D\}
\end{equation}
and hence $D \rightarrow L$ is deletable for each such $D$. \qed

\subsection{Intermediate Result: ``Add A Singleton Descendant to the Conditioning Set''}

The intuition behind the following lemma is that if $L$ is an
$\Hr$-lowest $\gleaf$, no v-structure below $L$ in $\Hr$ can be
``real'' in terms of the dependences in $\Gr$: for any $Y$ below $L$
that is independent of some other node $X$, they remain independent
when we condition on any singleton descendant $Z$ of $Y$, even if $Z$
is also a descendant of $X$. The lemma is stated in a somewhat
complicated manner because we want to use it both when (1) $X$ and $Y$
are adjacent but the edge is known to be deletable and (2) $X$ and $Y$
are not adjacent. We also find it convenient to include, in addition
to $Y$'s non-$X$ parents, an arbitrary additional set of
non-descendants $\bS$.

\begin{lemma} \label{lem:desc-add}
Let $Y$ be any $\Hr$-descendant of an $\Hr$-lowest $\gleaf$. If 
$$
\gcind{Y}{X}{\Parnot{Y}{X},\bS}
$$ for $\{X, \bS\} \subseteq \NonDesc{\Hr}{Y}$, then $ Y \indep_{\Gr}
X | \Parnot{Y}{X} , \bS , Z $ for any proper $\Hr$-descendant
$Z$ of $Y$.
\end{lemma}
{\noindent \bf Proof:} To simplify notation, let $\bR = \Parnot{Y}{X}
, \bS$.  Assume the lemma does not hold and thus $ Y
\notindep_{\Gr} X | \bR , Z $.  Consider any $(\bR , Z)$-active
path $\pi_{XY}$ between $X$ and $Y$ in $\Gr$. Because $Y \indep_{\Gr}
X | \bR$, this path cannot be active without $Z$ in the conditioning
set, which means that $Z$ must be on the path, and it must be a
collider in every position it occurs. Without loss of generality,
assume $Z$ occurs exactly once as a collider along the path (we can
simply delete the sub-path between the first and last occurrence of
$Z$, and the resulting path will remain active), and let $\pi_{XZ}$ be
the sub-path from $X$ to $Z$ along $\pi_{XY}$, and let $\pi_{ZY}$ be
the sub-path from $Z$ to $Y$ along $\pi_{XY}$.

Because $Z$ is a proper descendant of $Y$ in $\Hr$, and $Y$ is a
descendant of an $\Hr$-lowest $\gleaf$, we know $Z$ cannot be a
$\gleaf$, else it would be lower than $L$ in $\Hr$. That means that in
$\Gr$, there is a directed path $\pi_{ZL'} = Z \rightarrow \ldots
\rightarrow L'$ consisting of at least one edge from $Z$ to some
$\gleaf$. No node $T$ along this path can be in $\bR$, else we could
splice in the path $Z \rightarrow \ldots \rightarrow T \leftarrow
\ldots \leftarrow Z$ between $\pi_{XZ}$ and $\pi_{ZY}$, and the
resulting path would remain active without $Z$ in the conditioning
set. Note that this means that $L'$ cannot belong to $\Parnot{Y}{X}
\subseteq \bR$. Similarly, the path cannot reach $X$ or $Y$, else we
could combine this out-of-$Z$ path with $\pi_{ZY}$ or $\pi_{XZ}$,
respectively, to again find an $\bR$-active path between $X$ and $Y$.
We know that in $\Hr$, $L'$ must be a non-descendant of $Y$, else $L'$
would be a lower $\gleaf$ than $L$ in $\Hr$. Because $X \cup \bR$
contains all of $Y$'s parents and none of its descendants, and because
(as we noted) $L'$ cannot be in $X \cup \bR$, we know $\Hr$ contains
the independence $Y \indep_{\Hr} L' | X , \bR$. But we just argued
that the (directed) path $\pi_{ZL'}$ in $\Gr$ does not pass through
any of $X , Y , \bR$, which means that it constitutes an out-of $Z$
$(\bR , X)$-active path that can be combined with $\pi_{ZY}$ to
produce a $(\bR , X)$-active path between $Z$ and $L'$, yielding a
contradiction. \qed

\subsection{Intermediate Result: The ``Weak-Transitivity Deletion'' Lemma}

The next lemma considers a collider $X \rightarrow Z \leftarrow Y$ in
$\Hr$ where either there is no edge between $X$ and $Y$ (i.e., the
collider is a v-structure) or the edge is deletable. The lemma states
that if $X$ and $Y$ remain independent when conditioning on their
common child---where all the non-$\{X,Y,Z\}$ parents of all three
nodes are also in the conditioning set---then one of the two edges
must be deletable.

\begin{lemma} \label{lem:desc-del}
Let $X \rightarrow Z$ and $Y \rightarrow Z$ be two edges in $\Hr$. If
$ X \indep_{\Gr} Y | \Parnot{X}{Y} , \Parnot{Y}{X} ,
\Parnot{Z}{XY} $ and $ X \indep_{\Gr} Y | \Parnot{X}{Y} ,
\Parnot{Y}{X} , \Parnot{Z}{XY} , Z$ (i.e., $Z$ added to the
conditioning set), then at least one of the following must hold: $Z
\indep_{\Gr} X | \Parnot{Z}{X}$ or $Z \indep_{\Gr} Y | \Parnot{Z}{Y}$.
\end{lemma}
{\noindent \bf Proof:} Let $\bS = \{\Parnot{X}{Y} , \Parnot{Y}{X}\}
\setminus \Parnot{Z}{XY}$ be the (non-$X$ and non-$Y$) parents of $X$
and $Y$ that are not parents of $Z$, and let $\bR = \Parnot{Z}{XY}$ be
all of $Z$'s parents other than $X$ and $Y$. Using this notation, we
can re-write the two conditions of the lemma as:
\begin{equation} \label{eq:base-no-Z}
X \indep_{\Gr} Y | \bR , \bS
\end{equation}
and
\begin{equation} \label{eq:base-with-Z}
X \indep_{\Gr} Y | Z , \bR , \bS
\end{equation}
From the Weak Transitivity axiom we conclude from these two
independences that either $Z \indep_{\Gr} X | \bR , \bS$ or $Z
\indep_{\Gr} X | \bR , \bS$. Assume the first of these is true
\begin{equation} \label{eq:wt1}
Z \indep_{\Gr} X | \bR , \bS
\end{equation}
If we apply the Composition axiom to the independences in Equation \ref{eq:base-no-Z}
and Equation \ref{eq:wt1} we get $X \indep_{\Hr} Y , Z
| \bR , \bS$; applying the Weak Union axiom we can then pull $Y$ into the
conditioning set to get:
\begin{equation} \label{eq:wu-aio}
Z \indep_{\Hr} X | \{Y , \bR\} , \bS
\end{equation}
Because $\{Y , \bR\} , X$ is precisely the parent set of $Z$,
and because $\bS$ (i.e., the parents of $Z$'s parents) cannot contain
any descendant of $Z$, we know by the Markov conditions that
\begin{equation} \label{eq:nd-aio}
Z \indep_{\Hr} \bS | \{Y , \bR\} , X
\end{equation}
Applying the Intersection Axiom to the independences in Equation \ref{eq:wu-aio} and
Equation \ref{eq:nd-aio} yields:
$$
Z \indep_{\Hr} X | Y , \bR
$$
Because $Y , \bR =  \Parnot{Z}{X}$, this means the first
independence implied by the lemma follows.

If the second of the two independence facts that follow from Weak
Transitivity hold (i.e., if $Z \indep_{\Gr} X | \bR , \bS$), then a
completely parallel application of axioms leads to the second
independence implied by the lemma. \qed

\subsection{Intermediate Result: The ``Move Lower'' Lemma}

\comment{
The next lemma, Lemma \ref{lem:delete-lower}, is a generalization of
Corollary \ref{cor:recurse}, where $X$ and $Y$ need not be adjacent;
Corollary \ref{cor:recurse} 
will follow rather easily one we establish this
lemma, and the non-adjacent version of the result is also used to
prove 
Theorem \ref{th:main}.
}

\begin{lemma} \label{lem:delete-lower}
Let $Y$ be any $\Hr$-descendant of an $\Hr$-lowest $\gleaf$.  If there
exists an $X \in \NonDesc{\Hr}{Y}$ that has a common $\Hr$-descendant
with $Y$ and for which
$$ 
Y \indep_{\Gr} X | \Parnot{Y}{X}
$$ 
then there exists an edge $W \rightarrow Z$ that is deletable in
$\Hr$, where $Z$ is a proper $\Hr$-descendant of $Y$.
\end{lemma}
{\noindent \bf Proof:} Let $Z$ be the {\em highest} common descendant
of $Y$ and $X$, let $D_Y$ be the {\em lowest} descendant of $Y$ that
is a parent of $Z$, and let $D_X$ be any descendant of $X$ that is a
parent of $Z$. We know that either (1) $D_Y = Y$ and $D_X = X$ or (2) $D_Y$
and $D_X$ are not adjacent and have no directed path connecting them; if this were not the case, and $\Hr$
contained a path $D_Y \rightarrow \ldots \rightarrow D_X$ ($D_X
\rightarrow \ldots \rightarrow  D_Y$) then $D_X$
($D_Y$) would be a higher common descendant than $Z$. This means that
in either case (1) or in case (2), we have
\begin{equation} \label{eq:DE}
D_Y \indep_{\Gr} D_X | \Parnot{D_Y}{D_X}
\end{equation}
For case (1), this is given to us explicitly in the statement of the
lemma, and for case (2), $\Parnot{D_Y}{D_X} = \Parset{\Hr}{D_Y}$ and
thus the independence holds from the Markov conditions in $\Hr$
because $D_X$ is a non-descendant of $D_Y$. Because in both cases we
know there is no directed path from $D_Y$ to $D_X$, we know that all
of $\Parnot{D_X}{D_Y}$ are non-descendants of $D_Y$, and thus we can
add them (via Composition and Weak Union) to the conditioning set of
Equation \ref{eq:DE}:
\begin{equation} \label{eq:DE-X}
D_Y \indep_{\Gr} D_X | \Parnot{D_Y}{D_X} , \Parnot{D_X}{D_Y}
\end{equation}
For any $P_Z \in \Parnot{Z}{D_YD_X}$ (i.e., any parent of $Z$
excluding $D_Y$ and $D_X$), we know that $P_Z$ cannot be a descendant
of $D_Y$, else $P_Z$ would have been chosen instead of $D_Y$ as the
lowest descendant of $Y$ that is a parent of $Z$. Thus, we can yet
again add to the conditioning set (via Composition and Weak Union) to
get:
\begin{equation} \label{eq:DE-R}
D_Y \indep_{\Gr} D_X | \Parnot{D_Y}{D_X} , \Parnot{D_X}{D_Y} , \Parnot{Z}{D_YD_X}
\end{equation}
Because no member of the conditioning set in Equation \ref{eq:DE-R} is
a descendant of $D_Y$, and because $D_Y$, by virtue of being a
descendant of $Y$, must also be a descendant of the $\Hr$-lowest
$\gleaf$, we conclude from Lemma \ref{lem:desc-add} that for (proper
$\Hr$-descendant of $Y$) $Z$ we have:
\begin{equation} \label{eq:DE-R-Z}
D_Y \indep_{\Gr} D_X | \Parnot{D_Y}{D_X} , \Parnot{D_X}{D_Y} ,
\Parnot{Z}{D_YD_X} , Z
\end{equation}
Given Equation \ref{eq:DE-R} and Equation \ref{eq:DE-R-Z}, we can
apply Lemma \ref{lem:desc-del} and conclude either (1) $Z \indep_{\Gr}
D_Y | \Parnot{Z}{D_Y}$ and hence $D_Y \rightarrow Z$ is deletable in
$\Hr$ or (2) $Z \indep_{\Gr} D_X | \Parnot{Z}{D_X}$ and hence $D_X
\rightarrow Z$ is deletable in $\Hr$ \qed

\begin{corollary} \label{cor:yab}
Let $L$ be an $\Hr$-lowest $\gleaf$, and let $A$ be any $\Hr$-highest
child of $L$. If there exists an edge $A \leftarrow B$ in $\Hr$ for
which $L$ and $B$ are not adjacent, then either $L \rightarrow A$ or
$A \leftarrow B$ is deletable in $\Hr$.
\end{corollary}
{\noindent \bf Proof:} Because $L$ is equal to (and thus a descendant
of) an $\Hr$-lowest $\gleaf$, it satisfies the requirement for ``$Y$''
in the statement of Lemma \ref{lem:delete-lower}. Because $A$ is the
highest child of $L$, $B$ cannot be a descendant of $L$ and thus
satisfies the requirement of ``$X$'' in the statement of Lemma
\ref{lem:delete-lower}. From the proof of the lemma, if we choose $A$
to be the highest-common descendant (i.e., ``$Z$''), the corollary
follows by noting that because $A$ is the highest $\Hr$-child of $L$,
$L$ must be a lowest parent of $A$, and thus we can choose $D_Y = L$
$D_X = B$. \qed

\subsection{Intermediate Result: ``The Move-Down Corollary''}

\begin{corollary} \label{cor:recurse}
Let $X \rightarrow Y$ be any deletable edge within $\Hr$ for which $Y$
is a descendant of an $\Hr$-lowest $\gleaf$. Then there exists an edge
$Z \rightarrow W$ that is deletable in $\Hr$ for which $Z$ and $W$
have no common descendants.
\end{corollary}
{\noindent \bf Proof:} If $X$ and $Y$ have a common descendant, we
know from Lemma \ref{lem:delete-lower} that there must be another
deletable edge $Z \rightarrow W$ for which $W$ is a proper descendant
of $Y$, and thus $Z$ and $W$ satisfy the conditions for ``$X$'' and
``$Y$'', respectively, in the statement of Lemma
\ref{lem:delete-lower}, but with a lower ``$Y$'' than we had before.
Because $\Hr$ is acyclic, if we repeatedly apply this argument we must
reach some edge for which the endpoints have no common
descendants. \qed

\subsection{Main Result: Proof of Theorem \ref{th:main}}

{\noindent \bf Theorem~\ref{th:main}}
{\em
If $\Gr < \cpdag$ for CPDAG $\cpdag$ and DAG $\Gr$, then for any EIH
property $\Pi$ that holds on $\Gr$, there exists a $\Pi$-consistent
$Delete(X,Y,\HSet)$ that when applied to $\cpdag$ results in the CPDAG
$\cpdag'$ for which $\Gr \leq \cpdag'$.
}

{\noindent \bf Proof:} Consider any DAG $\Hr^0$ in
$\EClass{\cpdag}$. From Theorem \ref{th:delete}, we know that there
exists either a covered edge or deletable edge in $\Hr^0$; if we
reverse any covered edge in DAG $\Hr^i$, the resulting DAG $\Hr^{i+1}$
(which is equivalent to $\Hr^i$) will be
closer to $\Gr$ in terms of total edge differences, and therefore
because $\Hr^0 \not = \Gr$ we must eventually reach an $\Hr = \Hr^i$
for which Theorem \ref{th:delete} identifies a deletable edge $e$. The
edge $e$ in $\Hr$ satisfies the preconditions of Corollary
\ref{cor:recurse}, and thus we know that there must also exist a
deletable edge $X \rightarrow Y$ in $\Hr$ for which $X$ and $Y$ have
no common descendants in $\ISG{\Hr}{\bV}$ for $\bV = \prunegh$.

Let $\Hr'$ be the DAG that results from deleting the edge $X
\rightarrow Y$ in $\Hr$. Because there is a GES delete operator
corresponding to every edge deletion in every DAG in
$\EClass{\cpdag}$, we know there must be a set $\HSet$ for which the
operator $Delete(X, Y, \HSet)$---when applied to $\cpdag$---results in
$\cpdag' = \EClass{\Hr'}$. Because $X \rightarrow Y$ is deletable in
$\Hr$, the operator satisfies the IMAP requirement in the theorem. For
the remainder of the proof, we demonstrate that it is $\Pi$-consistent.

Because all directed edges in $\cpdag'$ are compelled, these edges
must exist with the same orientation in all DAGs in
$\EClass{\cpdag'}$; it follows that any subset $\bW$ of the common
descendants of $X$ and $Y$ in $\cpdag'$ must also be common
descendants of $X$ and $Y$ in $\Hr'$. But because $X$ and $Y$ have no
common descendants in the ``pruned'' subgraph $\ISG{\Hr}{\bV}$, we
know that $\bW$ is contained entirely in the complement of $\bV$,
which means $\ISG{\Hr}{\bW} = \ISG{\Gr}{\bW}$; because $\Hr'$ is the same as
$\Hr$ except for the edge $X \rightarrow Y$, we conclude $\ISG{\Hr'}{\bW} =
\ISG{\Gr}{\bW}$.

We now consider the induced subgraph $\ISG{\Hr'}{\bW \cup X \cup Y}$
that we get by ``expanding'' the graph $\ISG{\Hr'}{\bW}$ to include
$X$ and $Y$. Because $X$ and $Y$ are not adjacent in $\Hr'$, and
because $\Hr'$ is acyclic, any edge in $\ISG{\Hr'}{\bW \cup X \cup Y}$
that is not in $\ISG{\Hr'}{\bW}$ must be directed from either $X$ or
$Y$ into a node in the descendant set $\bW$. Because all nodes in
$\bW$ are in the complement of $\bV$, these new edges must also exist
in $\Gr$, and we conclude $\ISG{\Hr'}{\bW \cup X \cup Y} =
\ISG{\Gr}{\bW \cup X \cup Y}$.  To complete the proof, we note that
because $\Pi$ is hereditary, it must hold on $\ISG{\Hr'}{\bW \cup X
  \cup Y}$. From Proposition \ref{prop:subeq}, we know $\ISG{\Hr'}{\bW
  \cup X \cup Y} \approx \ISG{\cpdag'}{\bW \cup X \cup Y})$, and
therefore because $\Pi$ is equivalence invariant, it holds for
$\ISG{\cpdag'}{\bW \cup X \cup Y}$. \qed


\begin{thebibliography}{10}

\bibitem{Abbeel2006}
Pieter Abbeel, Daphne Koller, and Andrew~Y. Ng.
\newblock Learning factor graphs in polynomial time and sample complexity.
\newblock {\em Journal of Machine Learning Research}, 7:1743--1788, 2006.

\bibitem{AMP97}
Steen~A. Andersson, David Madigan, and Michael~D. Perlman.
\newblock A characterization of {M}arkov equivalence classes for acyclic
  digraphs.
\newblock {\em Annals of Statistics}, 25:505--541, 1997.

\bibitem{BlairPeyton93}
Jean R.~S. Blair and Barry~W. Peyton.
\newblock An introduction to chordal graphs and clique trees.
\newblock In {\em Graph Theory and Sparse Matrix Computations}, pages 1--29,
  1993.

\bibitem{Chickering95uai}
David~Maxwell Chickering.
\newblock A transformational characterization of {Bayesian} network structures.
\newblock In S.~Hanks and P.~Besnard, editors, {\em Proceedings of the Eleventh
  Conference on Uncertainty in Artificial Intelligence, {\rm Montreal, QU}},
  pages 87--98. Morgan Kaufmann, August 1995.

\bibitem{Chickering96lns}
David~Maxwell Chickering.
\newblock Learning {Bayesian} networks is {NP}-{Complete}.
\newblock In D.~Fisher and H.J. Lenz, editors, {\em Learning from Data:
  Artificial Intelligence and Statistics V}, pages 121--130. Springer-Verlag,
  1996.

\bibitem{Chickering02JMLRb}
David~Maxwell Chickering.
\newblock Optimal structure identification with greedy search.
\newblock {\em Journal of Machine Learning Research}, 3:507--554, November
  2002.

\bibitem{CM02uai}
David~Maxwell Chickering and Christopher Meek.
\newblock Finding optimal {Bayesian} networks.
\newblock In A.~Darwiche and N.~Friedman, editors, {\em Proceedings of the
  Eighteenth Conference on Uncertainty in Artificial Intelligence, {\rm
  Edmonton, AB}}, pages 94--102. Morgan Kaufmann, August 2002.

\bibitem{ChickeringMeek2015UAI}
David~Maxwell Chickering and Christopher Meek.
\newblock Selective greedy equivalence search: Finding optimal {B}ayesian
  networks using a polynomial number of score evaluations.
\newblock In {\em Proceedings of the Thirty First Conference on Uncertainty in
  Artificial Intelligence, {\rm Amsterdam, Netherlands}}, 2015.

\bibitem{ChickeringMeek2015arxiv}
David~Maxwell Chickering and Christopher Meek.
\newblock Selective greedy equivalence search: Finding optimal {B}ayesian
  networks using a polynomial number of score evaluations.
\newblock 2015, arxiv:1506.2849v1.

\bibitem{CMH2004}
David~Maxwell Chickering, Christopher Meek, and David Heckerman.
\newblock Large-sample learning of {B}ayesian networks is {NP}-hard.
\newblock {\em Journal of Machine Learning Research}, 5:1287--1330, October
  2004.

\bibitem{Chow68}
C.~Chow and C.~Liu.
\newblock Approximating discrete probability distributions with dependence
  trees.
\newblock {\em IEEE Transactions on Information Theory}, 14:462--467, 1968.

\bibitem{Dasgupta99}
S.~Dasgupta.
\newblock Learning polytrees.
\newblock In K.~Laskey and H.~Prade, editors, {\em Proceedings of the Fifteenth
  Conference on Uncertainty in Artificial Intelligence, {\rm Stockholm,
  Sweden}}, pages 131--141. Morgan Kaufmann, 1999.

\bibitem{Edmonds67}
J.~Edmonds.
\newblock Optimum branching.
\newblock {\em J. Res. NBS}, 71B:233--240, 1967.

\bibitem{Friedman1999uai}
Nir Friedman, Iftach Nachman, and Dana Peer.
\newblock Learning bayesian network structure from massive datasets: The
  ``sparse candidate'' algorithm.
\newblock In {\em Proceedings of the Fifteenth Conference on Uncertainty in
  Artificial Intelligence, {\rm Stockholm, Sweden}}. Morgan Kaufmann, 1999.

\bibitem{Gaspers2012}
Serge Gaspers, Mikko Koivisto, Mathieu Liedloff, Sebastian Ordyniak, and Stefan
  Szeider.
\newblock On finding optimal polytrees.
\newblock In {\em Proceedings of the Twenty-Sixth AAAI Conference on Artificial
  Intelligence}. AAAI Press, 2012.

\bibitem{GPP1990}
Dan Geiger, Azaria Paz, and Judea Pearl.
\newblock Learning causal trees from dependence information.
\newblock In {\em Proceedings of the Eighth National Conference on Artificial
  Intelligence - Volume 2}, AAAI'90, pages 770--776. AAAI Press, 1990.

\bibitem{GP01}
Steven~B. Gillispie and Michael~D. Perlman.
\newblock Enumerating {M}arkov equivalence classes of acyclic digraph models.
\newblock In M.~Goldszmidt, J.~Breese, and D.~Koller, editors, {\em Proceedings
  of the Seventeenth Conference on Uncertainty in Artificial Intelligence, {\rm
  Seattle, WA}}, pages 171--177. Morgan Kaufmann, 2001.

\bibitem{Kalisch2007}
Markus Kalisch and Peter Buhlmann.
\newblock Estimating high-dimensional directed acyclic graphs with the pc
  algorithm.
\newblock {\em Journal of Machine Learning Research}, 8:613--636, 2007.

\bibitem{KargerSrebro2001}
David Karger and Nathan Srebro.
\newblock Learning {M}arkov networks: Maximum bounded tree-width graphs.
\newblock In {\em 12th ACM-SIAM Symposium on Discrete Algorithms (SODA)}, pages
  391--401, January 2001.

\bibitem{KoivistoSood2004}
Mikko Koivisto and Kismat Sood.
\newblock Exact bayesian structure discovery in bayesian networks.
\newblock {\em J. Mach. Learn. Res.}, 5:549--573, December 2004.

\bibitem{Kojima2010}
Kaname Kojima, Eric Perrier, Seiya Imoto, and Satoru Miyano.
\newblock Optimal search on clustered structural constraint for learning
  {B}ayesian network structure.
\newblock {\em Journal of Machine Learning Resarch}, 11:285--310, 2010.

\bibitem{Meek2001}
Christopher Meek.
\newblock Finding a path is harder than finding a tree.
\newblock {\em Journal of Artificial Intelligence Research}, 15:383--389, 2001.

\bibitem{NarasimhanBilmes2004uai}
Mukund Narasimhan and Jeff Bilmes.
\newblock Pac-learning bounded tree-width graphical models.
\newblock In {\em Proceedings of the 20th Conference on Uncertainty in
  Artificial Intelligence}, UAI '04, pages 410--417, Arlington, Virginia,
  United States, 2004. AUAI Press.

\bibitem{Ordyniak2013}
Sebastian Ordyniak and Stefan Szeider.
\newblock Parameterized complexity results for exact bayesian network structure
  learning.
\newblock {\em Journal of Artificial Intelligence Research}, 46:263--302, 2013.

\bibitem{Pearl88}
Judea Pearl.
\newblock {\em Probabilistic Reasoning in Intelligent Systems: Networks of
  Plausible Inference}.
\newblock Morgan Kaufmann, San Mateo, CA, 1988.

\bibitem{Shahaf2009}
Dafna Shahaf, Anton Chechetka, and Carlos Guestrin.
\newblock Learning thin junction trees via graph cuts.
\newblock In {\em In Artificial Intelligence and Statistics (AISTATS)},
  Clearwater Beach, Florida, April 2009.

\bibitem{Silander2006}
Tomi Silander and Petri Myllym\"{a}ki.
\newblock A simple approach for finding the globally optimal bayesian network
  structure.
\newblock In {\em Proceedings of the Twenty Second Conference on Uncertainty in
  Artificial Intelligence, {\rm Cambridge, MA}}, pages 445--452, 2006.

\bibitem{Spirtes93}
Peter Spirtes, Clark Glymour, and Richard Scheines.
\newblock {\em Causation, Prediction, and Search}.
\newblock Springer-Verlag, New York, 1993.

\bibitem{Tsamardinos2006ML}
Ioannis Tsamardinos, Laura~E. Brown, and Constantin~F. Aliferis.
\newblock The max-min hill-climbing {B}ayesian network structure learning
  algorithm.
\newblock {\em Machine Learning}, 2006.

\bibitem{VP91}
Thomas Verma and Judea Pearl.
\newblock Equivalence and synthesis of causal models.
\newblock In M.~Henrion, R.~Shachter, L.~Kanal, and J.~Lemmer, editors, {\em
  Proceedings of the Sixth Conference on Uncertainty in Artificial
  Intelligence}, pages 220--227, 1991.

\end{thebibliography}
\end{document}